\pdfoutput=1

\documentclass[11pt]{article}

\usepackage[final]{acl}

\usepackage{times}
\usepackage{latexsym}
\usepackage[utf8]{inputenc} 
\usepackage[T1]{fontenc}    
\usepackage{hyperref}       
\usepackage{url}            
\usepackage{booktabs}       
\usepackage{amsfonts}       
\usepackage{nicefrac}       
\usepackage{microtype}      
\usepackage{xcolor}         
\usepackage{graphicx}
\usepackage{amsmath}
\usepackage{amsfonts}
\usepackage{multirow}
\usepackage{subfigure}
\usepackage{subcaption}
\usepackage{pgfplots}
\usepackage{xspace}
\usepackage{enumitem}
\usepackage{wrapfig}
\usepackage{bbding}
\usepackage{colortbl}
\usepackage{adjustbox}
\usepackage{float}
\usepackage{tabularx}

\usepackage[T1]{fontenc}

\usepackage[utf8]{inputenc}

\usepackage{microtype}

\usepackage{inconsolata}

\usepackage{graphicx}

\definecolor{mycolor_green}{HTML}{D5E8D4}
\definecolor{mycolor_orange}{HTML}{FFE6CC}
\definecolor{mycolor_blue}{HTML}{DAE8FC}
\definecolor{mycolor_red}{HTML}{F8CECC}

\newcommand*\samethanks[1][\value{footnote}]{\footnotemark[#1]}

\title{Chain of Strategy Optimization Makes \\ Large Language Models Better Emotional Supporter}

\author{Weixiang Zhao$^1$\thanks{\ \ \ Equal contribution}, Xingyu Sui$^1$\samethanks, Xinyang Han$^1$, Yang Deng$^2$, Yulin Hu$^1$,  Jiahe Guo$^1$ \\ \textbf{Libo Qin}$^3$, \textbf{Qianyun Du}$^4$, \textbf{Shijin Wang}$^4$, \textbf{Yanyan Zhao}$^1$\thanks{\ \ Corresponding author}, \textbf{Bing Qin}$^1$, \textbf{Ting Liu}$^1$ \\
        $^1$Harbin Institute of Technology,
        $^2$Singapore Management University \\ $^3$Central South University, $^4$iFLYTEK AI Research (Central China), iFLYTEK Co., Ltd\\
        \texttt{\{wxzhao, xysui, yyzhao\}@ir.hit.edu.cn}}

\begin{document}
\maketitle
\begin{abstract}
The growing emotional stress in modern society has increased the demand for Emotional Support Conversations (ESC). While Large Language Models (LLMs) show promise for ESC, they face two key challenges: (1) low strategy selection accuracy, and (2) preference bias, limiting their adaptability to users’ emotional needs. Existing supervised fine-tuning (SFT) struggles to address these issues, as it rigidly trains models on single gold-standard responses without modeling nuanced strategy trade-offs. To overcome these limitations, we propose a novel two-stage framework that optimizes strategy selection preferences at each dialogue turn. We first leverage Monte Carlo Tree Search to construct ESC-Pro, a high-quality preference dataset with turn-level strategy-response pairs. Then training on ESC-Pro with Chain-of-Strategy Optimization (CSO) improves both strategy accuracy and bias mitigation, enabling LLMs to generate more empathetic and contextually appropriate responses. Experiments on LLaMA-3.1-8B, Gemma-2-9B, and Qwen2.5-7B demonstrate that CSO outperforms standard SFT, highlighting the efficacy of fine-grained, turn-level preference modeling in ESC.\footnote{Our code and data can be found in \url{https://github.com/XingYuSSS/CSO}.}
\end{abstract}

\section{Introduction}

In modern society, people increasingly face emotional stress due to mounting work and life pressures. As a result, the demand for Emotional Support Conversations (ESC) has grown significantly, providing individuals with psychological relief and guidance \citep{langford1997social,greene2003handbook,heaney2008social}. High-quality ESC can help alleviate emotional distress, offering comfort and constructive advice \citep{burleson2003emotional}. With the rapid advancement of large language models (LLMs) \citep{brown2020language,dubey2024llama,team2024gemma,yang2024qwen2}, their exceptional conversational abilities have opened up new possibilities for ESC.

\begin{figure}
    \centering
    \includegraphics[width=1\linewidth]{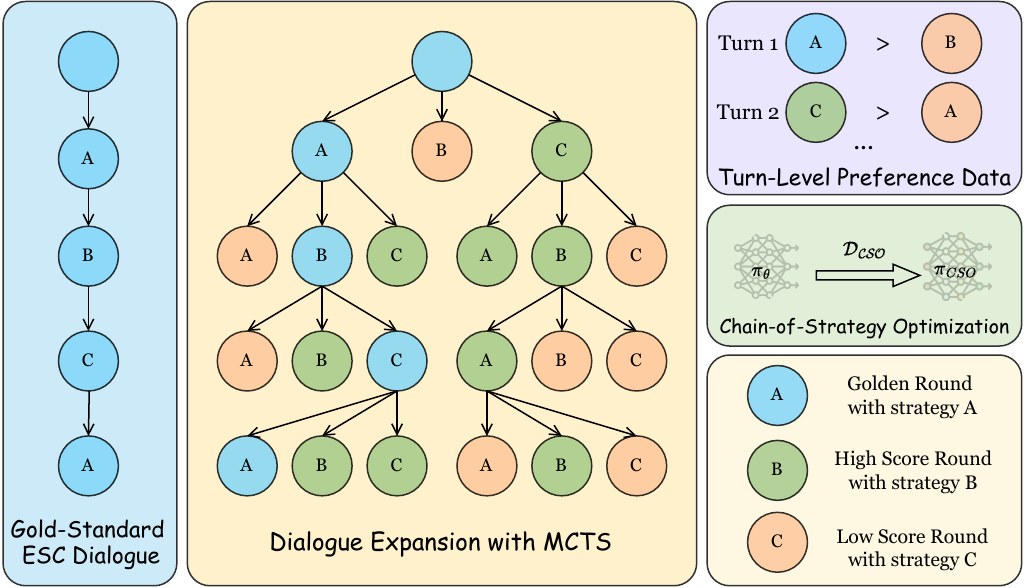}
    \caption{Left: Supervised fine-tuning on glod-standard conversation. Middle: Expanding existing conversations with MCTS. Right: Constructing preference dataset and conducting preference optimization.}
    \label{fig:example}
    \vspace{-3mm}
\end{figure}

However, achieving high-quality ESC with LLMs remains a significant challenge due to two core limitations: 1) LLMs struggle with \textbf{low strategy selection accuracy}, often failing to choose contextually appropriate support strategies \citep{zhao2023chatgpt,chen2023soulchat,farhat2024chatgpt}. 2) LLMs exhibit \textbf{strategy preference bias}, favoring certain strategies rigidly rather than adapting to users’ dynamic emotional needs \citep{kang2024large}. Due to the strategy-driven nature of ESC, nuanced trade-offs between strategies are critical \citep{liu2021towards,deng2023knowledge,zhao2023transesc}. While supervised fine-tuning (SFT) is the mainstream training approach \citep{ouyang2022training,zheng2024self}, its inherent rigidity exacerbates these challenges. By training exclusively on single golden strategies and responses, SFT teaches LLMs which strategies to apply but fails to clarify why certain strategies are inappropriate in specific contexts. This leaves models unable to grasp the contextual nuances required for dynamic adaptation.

To tackle these challenges, we advocate for a more fine-grained and turn-level approach to optimizing ESC. Specifically, we propose a two-stage framework that first constructs rich preference data at the (strategy, response) level, and then refines model behavior through preference learning.


In the first stage, we utilize Monte Carlo Tree Search (MCTS) to expand gold-standard ESC conversations into a conversation tree, where each layer represents a dialogue turn, and sibling nodes correspond to different strategic responses. As illustrated in the middle of Figure~\ref{fig:example}, we design a comprehensive value function that evaluates each response along four key dimensions—Empathy, Information, Humanoid, and Strategy—to ensure high-quality exploration. From this tree, we extract a refined \underline{\textbf{ESC}} dataset for \underline{\textbf{Pr}}eference \underline{\textbf{o}}ptimization, \textbf{ESC-Pro}, as shown in the right of Figure \ref{fig:example}. Specifically, low-scoring nodes are categorized as non-prefer samples, identifying suboptimal strategies, while both the original gold-standard nodes and newly discovered high-scoring nodes are included as prefer samples, enhancing the dataset with high-quality strategy examples.

In the second stage, we introduce \underline{\textbf{C}}hain-of-\underline{\textbf{S}}trategy \underline{\textbf{O}}ptimization (\textbf{CSO}), a preference optimization approach that explicitly targets strategy-level improvements. As shown in the right of Figure~\ref{fig:example}, training on ESC-Pro with CSO enables LLMs to not only select better strategies over weaker ones at each dialogue turn, but also to explore diverse conversation paths and avoid rigid preference patterns. This chain-like optimization across turns improves both local adaptability and long-range strategic coherence.

Our comprehensive experiments on LLaMA-3.1-8B \citep{dubey2024llama}, Gemma-2-9B \citep{team2024gemma}, and Qwen2.5-7B \citep{yang2024qwen2} demonstrate that CSO is highly effective in raising the accuracy of strategy selection and mitigating strategy preference bias. To further validate CSO, we instantiated multiple preference optimization methods \citep{hong2024orpo,meng2024simpo} on ESC-Pro and consistently observed superior performance compared to standard SFT. This consistent improvement underscores the effectiveness of ESC-Pro as a high-quality dataset and highlights the importance of fine-grained, turn-level preference modeling in achieving effective ESC.

The main contributions of this work are summarized as follows: (1) We present ESC-Pro, a high-quality turn-level preference dataset for ESC, constructed using MCTS. (2) We propose CSO, enabling LLMs to learn nuanced strategy trade-offs at each dialogue turn. (3) Extensive experiments on three LLMs demonstrate CSO is effective in improving both ESC quality and adaptability.

\section{Related Work}

\paragraph{Emotional Support Conversation}

Emotional Support Conversations (ESC) \cite{liu2021towards} center around interactions between a user, referred to as the seeker, who is experiencing emotional distress, and a supporter, whose goal is to alleviate the seeker’s emotional intensity by utilizing a set of strategies to guide the conversation. Various approaches have been proposed to build the ESC systems, such as global-to-local hierarchical graph network \cite{peng2022control}, incorporating commonsense knowledge \cite{tu2022misc}, and modeling emotions and semantics \cite{zhao2023don,zhao2023transesc}. With the development of LLMs, some works aim to directly leverage the performance of these models without altering their architecture. \citet{liu2023chatcounselor} apply SFT to the LLaMA-7B model for the ESC task and introduce ChatCounselor, a model specialized in ESC tasks that outperforms general-purpose models. \citet{chen2023soulchat} and \citet{qiu2023smile} expand single-turn empathic responses into multi-turn dialogues and performed fine-tuning on a high-quality ESC dataset they constructed, thereby improving the model’s ESC performance.

\begin{figure*}
    \centering
    \includegraphics[width=1\linewidth]{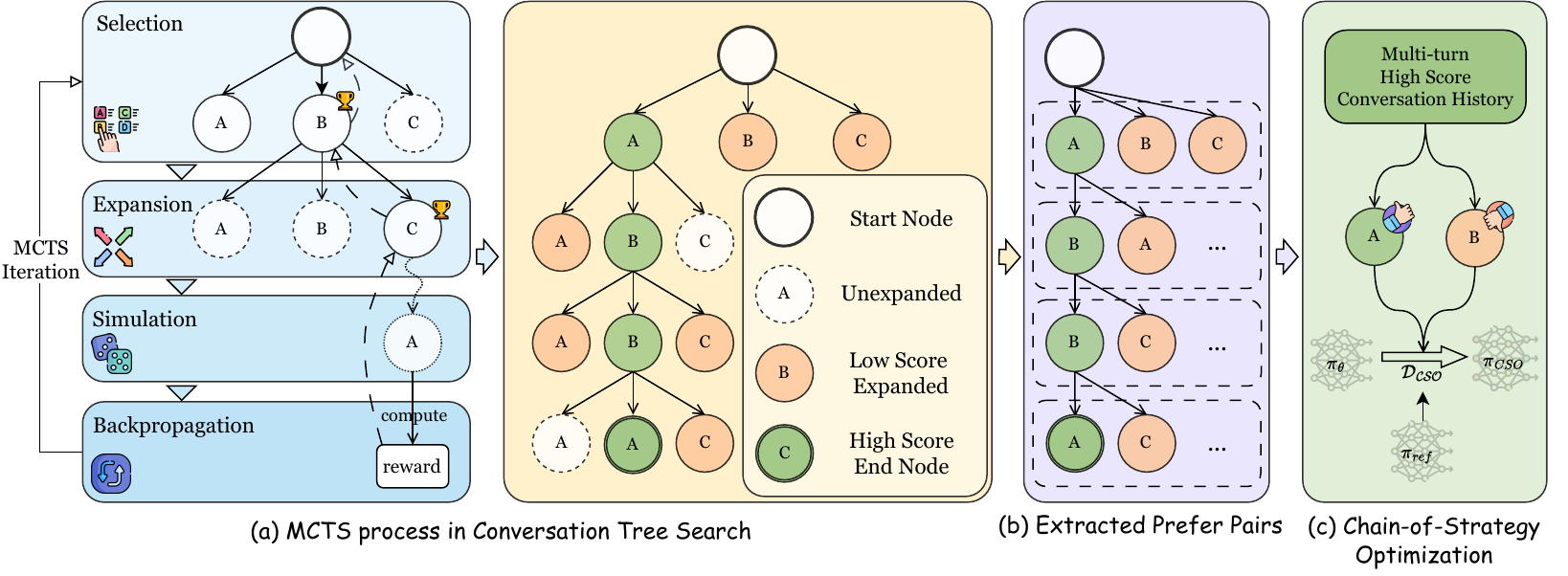}
    \caption{The overall framework. (a) Dialogue Expansion with MCTS: We leverage Monte Carlo Tree Search (MCTS) to systematically explore and refine ESC strategies by constructing a structured conversation tree. (b) Preference Data Construction: We extract high-quality strategy preference pairs from the expanded conversation tree to form the ESC-Pro dataset. (c) Chain-of-Strategy Optimization (CSO): We fine-tune LLMs on ESC-Pro using preference optimization techniques to enhance strategy selection accuracy and mitigate preference bias.}
    \label{fig:framework}
    \vspace{-3mm}
\end{figure*}

\paragraph{Preference Optimization Algorithms}

Preference optimization algorithms align model outputs with user preferences by training on pairs of positive and negative samples, enhancing the fine-tuning of LLMs. Traditional approaches use policy-based reinforcement learning (RL) to train reward models \citep{ouyang2022training}, followed by online RL methods like Proximal Policy Optimization (PPO) \citep{schulman2017proximal} for LLM fine-tuning. The Direct Preference Optimization (DPO) algorithm \citep{rafailov2023direct} streamlined this process by unifying reward modeling and RL into a single step, reducing computational costs. Subsequent methods further optimized DPO: SimPO \citep{meng2024simpo} eliminates DPO’s reference model, IPO \citep{azar2024general} enhances robustness via regularization, ORPO \citep{hong2024orpo} integrates instruction tuning and preference alignment, and KTO \citep{ethayarajh2024kto} operates without paired data by leveraging prospect theory. These advancements simplify training, reduce resource demands, and broaden data applicability.

Existing ESC models rely on SFT, which lacks the ability to differentiate between optimal and suboptimal strategies. Our work bridges this gap by applying preference optimization to ESC, enhancing strategy accuracy and reducing bias.

\section{Methodology}
We introduce a novel framework that refines strategy selection at each dialogue turn through structured preference modeling. As shown in Figure \ref{fig:framework}, this optimization paradigm consists of three key components: (1) Dialogue Expansion with MCTS, (2) Preference Data Construction and (3) Chain-of-Strategy Optimization (CSO).

\subsection{Dialogue Expansion with MCTS}

To enrich exists ESC dataset with turn-level preference annotation, we employ Monte Carlo Tree Search (MCTS) to construct a structured conversation tree, where each node represents a specific dialogue state. The search process iteratively refines dialogue strategies through four key stages: selection, expansion, simulation, and backpropagation. This enables the model to explore various strategy paths and identify optimal responses based on a predefined value function.

To guide the dialogue generation process, we define four specialized LLMs, each playing a distinct role in constructing and evaluating the conversation tree:
\textbf{Seeker LLM}: Generates responses based on the conversation history from the seeker perspective.
\textbf{Supporter LLM}: Produces replies based on the conversation history and a given strategy from the supporter perspective.
\textbf{Strategy LLM}: Evaluates and ranks available support strategies for each conversation turn.
\textbf{Reward LLM}: Assesses the quality of each strategy-response pair with four key metrics: empathy, information, humanoid quality, and strategy effectiveness, serving as the value function for MCTS. The detailed setups for these LLMs are provided in Appendix \ref{sec:llm_def}.

\subsubsection{Node Types and Representation}

Each node in the conversation tree represents a state, classified into four types:

\paragraph{Root Node} The root node represents the initial dialogue turn, containing the Seeker’s first response $R_{\text{seeker}_0}$, Q-value $Q$, and visit count $N$:
\begin{align}
S_{\text{root}} = (R_{\text{seeker}_0}, Q, N) \notag
\end{align}

\paragraph{Unexpanded Node} An unexpanded node represents an unexplored strategy, with the strategy $a$, score $r$, Q-value (initialized to 0), and visit count (initialized to 0):
\[
S_{\text{unexp}} = (
a, r, Q = 0, N = 0)
\]

\paragraph{Expanded Node} An expanded node has been explored, including the strategy, score, Seeker and Supporter 
responses, Q-value, and visit count:
\[
S_{\text{exp}} = (
a, r, R_{\text{seeker}}, R_{\text{supporter}}, Q, N)
\]

\paragraph{End Node} An end node marks the end of the dialogue, including the strategy, score, Supporter's 
response, Q-value, and visit count:
\[
S_{\text{end}} = (
a, r, R_{\text{supporter}}, Q, N)
\]

\subsubsection{Selection}

During the selection phase, we identify the next node to explore using the Polymer Upper Confidence Bound (PUCB) \citep{rosin2011multi} formula. This formula balances exploration and exploitation based on the node's Q-value, visit count, and parent node’s visit count. The formula is:
\begin{equation}\small
\text{PUCB}(S) = Q(S) + c \cdot P(S) \cdot \frac{\sqrt{N(\text{Parent}(S))}}{N(S) + 1} 
\label{eq:pucb}
\end{equation}
where $c$ is a hyper-parameter. The selection process proceeds layer by layer, starting from the root node and selecting the node with the highest PUCB value at each step.

\subsubsection{Expansion}

In the expansion phase, the selected node is evaluated based on its type. If the node is an \textbf{expanded node}, the Strategy LLM evaluates all possible strategies for the Supporter at this node, assigns scores, and generates unexpanded child nodes for each strategy. These child nodes are initialized with Q and N values set to 0:
\[
S_{\text{child}} = (
a, r, Q = 0, N = 0)
\]

If the selected node is an \textbf{unexpanded node}, the conversation history is generated using the Seeker and Supporter LLMs: (1) First, the Supporter LLM generates a response based on the selected strategy and the conversation history. (2) Next, the Seeker LLM generates a reply based on the conversation history and the Supporter’s response.

Once the node is expanded, the conversation history is stored in the node. If the Seeker generates an end-of-dialogue token during this process, the reward is immediately computed, and the process moves to backpropagation.

\subsubsection{Simulation}

The simulation focuses on the newly expanded child nodes. To reduce computational costs, only the child node with the highest strategy score is selected for simulation. This proceeds as follows:

\paragraph{Node Expansion} The chosen child node is expanded from an unexpanded node to an expanded node. The Seeker and Supporter LLMs generate the conversation content as needed.

\paragraph{Greedy Simulation} A greedy simulation is performed for \( n \) steps, where at each step: (1) The Seeker LLM generates a response based on the highest-scoring strategy. (2) The Supporter LLM responds accordingly.

The simulation continues for \( n \) steps or until an end-of-dialogue token is generated by the Seeker.

\paragraph{Reward Calculation} During the simulation, the Reward LLM evaluates the conversation quality using four metrics: Empathy (\( E \)), Information (\( I \)), Humanoid (\( H \)), and Strategy (\( S \)). The reward is computed as:
\begin{equation}\small
R = \frac{E + I + H + \alpha \cdot S}{10} + b
\label{eq:reward_calc}
\end{equation}
where \( E \), \( I \), \( H \), and \( S \) are calculated based on the conversation history at each step. $\alpha$ is a scaling hyper-parameter. A bias \( b \) is introduced to adjust the reward, allows the system to treat rewards lower than \( -b \) as negative, helping guide the search towards higher-scoring nodes. The reward for the simulation is averaged over all Supporter turns during the simulation and is used to update the node’s Q-value in the backpropagation phase:
\begin{equation}\small
R_{\text{sim}}(S) = \frac{1}{r_{\text{sim\_end}} - r(S)} \sum\nolimits_{i={r(S)}}^{r_{\text{sim\_end}}} R_{\text{sim}}^{(i)}
\end{equation}

\subsubsection{Backpropagation}

In the backpropagation phase, the reward \( R_{\text{sim}}(S) \) is propagated backward from the simulated node. The Q-value and visit count for each node are updated as follows:
\begin{equation}\small
Q_k = \frac{N_k \cdot Q_k + R_{\text{sim}}}{N_k + 1}, \quad N_k = N_k + 1
\end{equation}

This process updates the tree and refines the search, improving future strategy selections. By iterating through these four stages, the MCTS process efficiently optimizes the dialogue strategy, balancing exploration and exploitation, while utilizing the LLMs to guide the conversation and evaluate strategies based on rewards.

\subsection{ESC-Pro Preference Data Construction}
Based on MCTS-based dialogue expansion, we construct the ESC-Pro dataset with strategy preferences at each dialogue turn. This dataset is derived from the conversation tree by identifying high-quality strategy-response paths and pairing them with lower-scoring alternatives to create fine-grained preference data.

\paragraph{Conversation Decomposition and Expansion}

To construct the preference dataset, we decompose a gold-standard ESC conversation into a structured conversation tree, where each layer corresponds to a specific dialogue turn. Sibling nodes within a layer represent different strategic choices. The tree expands iteratively through MCTS-based search, ensuring comprehensive exploration of potential strategy paths while maintaining computational efficiency. The process stops when: The search reaches a predefined number of iterations \( n_{\text{iter}} \) or a sufficient number of termination nodes \( n_{\text{finish}} \) have been identified.

\paragraph{Preference Data Extraction}

After completing the MCTS process, we extract valid conversation paths from the tree, where each node in the path satisfies the condition \( Q(S_i) > \theta \). A path \( P = \{S_1, S_2, \dots, S_L\} \) is valid if:
\begin{equation}
\text{is\_end\_node}(S_L) \ \; \text{and} \ \;  Q(S_i) > \theta \ \;  \forall S_i \in P
\label{eq:path}
\end{equation}

Here, \( \text{is\_end\_node}(S_L) \) ensures that the last node in the path, \( S_L \), is a termination node, and \( Q(S_i) > \theta \) ensures that all nodes in the path meet the quality threshold. 

Once valid paths are identified, we extract preference pairs by identifying low-scoring siblings \( S_l \) for each high-scoring node \( S_w \), where \( Q(S_l) < \theta \). These pairs \( (S_w, S_l) \) represent relative strategy quality and are used to train the model. 

The resulting dataset, denoted as \( \mathcal{D} \), is constructed as follows:
\begin{equation}\small
    \mathcal{D} = \bigcup_{P \in \mathcal{P}} \left\{
    \begin{aligned}
    & \{ (S_w, S_l) \mid S_w \in P, S_l \in \text{Siblings}(S_w), \\
    & \quad Q(S_w) > \theta, Q(S_l) < \theta \}
    \end{aligned}
    \right.
    \label{eq:dataset}
\end{equation}

where \( \mathcal{P} \) denotes the set of all valid paths, \( \text{siblings}(S_w) \) denotes all sibling nodes of the node \( S_w \). The dataset \( \mathcal{D} \) contains all preference pairs \( (S_w, S_l) \) extracted from valid paths. By incorporating both preferred and non-preferred strategies, ESC-Pro provides a rich training signal, allowing LLMs to learn nuanced strategy trade-offs and improve adaptive decision-making.

\subsection{Chain-of-Strategy Optimization}

We perform turn-level preference optimization with DPO \citep{rafailov2023direct} on the ESC-Pro dataset \( \mathcal{D} \). For the \(i\)-th conversation round, the training objective is formulated as follows:
\begin{equation}\small
\mathcal{L}_{i}(\pi_{\theta}; \pi_{\text{ref}}) = - \log \sigma \left( \beta \log r_w - \beta \log r_l \right)
\end{equation}
\begin{equation}\small
r_w = \frac{\pi_{\theta}(S^i_w \mid x, H^{i-1})}{\pi_{\text{ref}}(S^i_w \mid x, H^{i-1})}, r_l = \frac{\pi_{\theta}(S^i_l \mid x, H^{i-1})}{\pi_{\text{ref}}(S^i_l \mid x, H^{i-1})}
\end{equation}
where \(H^{i-1}\) represents \(\{S_w^0, S_w^1, \dots, S_w^{i-1}\}\). The overall training objective is:
\begin{equation}\small
\mathcal{L}_{\text{CSO}}(\pi_{\theta}; \pi_{\text{ref}}) = - \mathbb{E}_{(x, S_w^i, S^i_l, H^{i-1}) \sim D} \left[ \mathcal{L}_{i}(\pi_{\theta}; \pi_{\text{ref}})\right]
\end{equation}

\section{Dataset Quality}

\subsection{Statics of ESC-Pro}

We expand 100 seed dialogues from ExTES \citep{zheng2024self} into 423 dialogues, forming our ESC-Pro dataset. The total number of utterances grows from 1,613 to 14,383, with over half (8,157 utterances) classified as non-preference data. This demonstrates that our method not only effectively expands high-quality preference data but also generates a substantial amount of non-preference data, making ESC-Pro well-suited for preference optimization. Please refer to Appendix \ref{app:data_statics} for detailed results and discussion.

The average dialogue length remains consistent between the expanded dataset (14.72 utterances) and the original (16.13 utterances), ensuring that expansion does not degrade data quality. Additionally, the average length of preference utterances (29.42) closely matches that of the seed data (29.03), while non-preference utterances (23.22) are notably shorter. This distinction highlights the effectiveness of our method in capturing meaningful preference differences within ESC interactions.

\subsection{Data Quality Evaluation}

\begin{table}
\centering
\resizebox{\linewidth}{!}{
\begin{tabular}{@{}ccccc@{}}
\toprule
{ESC-Pro(+) vs. ESC-Pro(-)} & Win & Lose & Tie & $\kappa$ \\
\midrule
Empathy & \textbf{46.33} & 32.67 & 21.00 & 0.61 \\
Information & \textbf{42.34} & 27.33 & 30.33 & 0.55 \\
Humanoid & \textbf{41.67} & 21.33 & 37.00 & 0.49 \\
Strategy & \textbf{60.67} & 15.00 & 24.33 & 0.67 \\
\bottomrule
\end{tabular}
}
\caption{Pairwise comparison results between ESC-Pro(+) and ESC-Pro(-). The ``Win'' column indicates cases where the preference response is rated higher, while ``Lose'' represents cases where the non-preference response was preferred. The $\kappa$ coefficient measures inter-rater agreement.}
\label{tab:preference_pair_comparison}
\end{table}

To assess the quality gap between preference and non-preference data, we conduct a pairwise comparison using the four evaluation metrics from the Reward LLM: Empathy, Information, Humanoid, and Strategy. Evaluators compare 100 preference pairs from ESC-Pro and determine whether the preference response is superior, inferior, or equal to the non-preference response. As shown in Table \ref{tab:preference_pair_comparison}, preference data consistently outperforms non-preference data, particularly in Strategy (winning in 61 cases vs. 15 losses), aligning with the Reward LLM’s weighting scheme. The Empathy and Information metrics also favor preference data, while the Humanoid metric shows a more balanced distribution. These results confirm both the effectiveness of ESC-Pro in capturing high-quality strategy responses and the reliability of the Reward LLM’s scoring methodology. Please refer to Appendix \ref{app:data_quality} for more results on the data quality evaluation.

\begin{figure}
    \centering
    \includegraphics[width=1\linewidth]{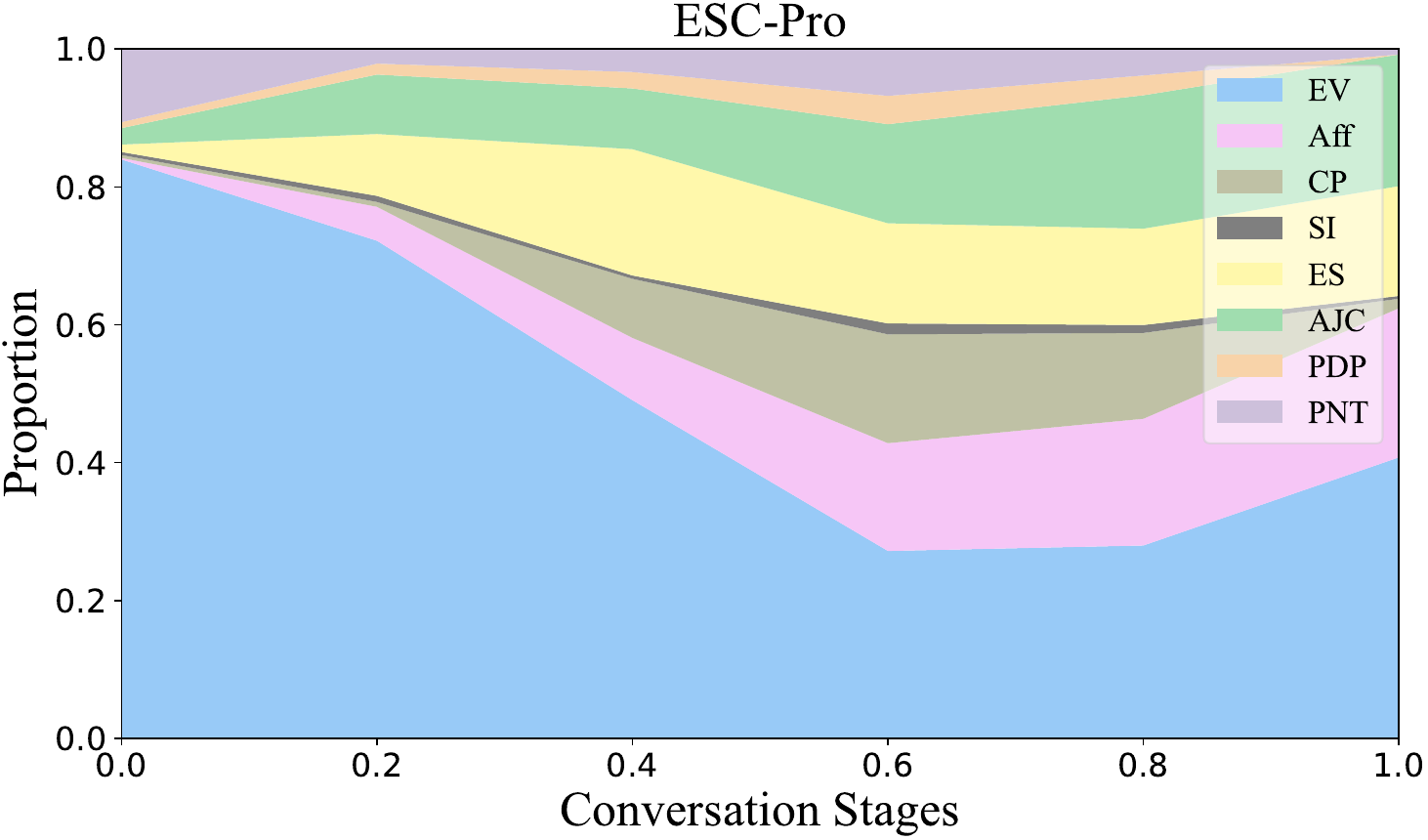}
    \caption{Strategy distribution across conversation stages in ESC-Pro.}
    \label{fig:strategy_distribution}
\end{figure}

\subsection{Strategy Analysis}

The ExTES dataset originally contains 16 distinct support strategies. To reduce the search space and improve computational efficiency, we merge similar strategies into 8 categories (see Appendix \ref{app:strategies_def} for details). We analyze the distribution of these strategies across six conversation stages. Given a dialogue with $N$ utterances, each utterance $k$ (where $k = 1, …, N$) is assigned to stage $i$ using:
\[
i = \left\lfloor \frac{k}{N} \times 6 \right\rfloor \times 0.2
\]
where $i$ ranges from 0 to 1 in increments of 0.2.

As shown in Figure \ref{fig:strategy_distribution}, the ESC-Pro dataset exhibits a dynamic and contextually appropriate strategy distribution. And the comparison with the seed dataset reveals that ESC-Pro employs a more diverse and balanced strategy distribution. Please refer to Appendix \ref{app:strategy} for more details.

\subsection{Toxicity Analysis}

We assess the toxicity levels of ESC-Pro using the Perspective API, a widely used tool for detecting harmful content. Our analysis shows that ESC-Pro maintains a similar toxicity profile to ExTES-seed, ensuring its suitability for preference optimization. Please refer to Appendix \ref{app:toxicity} for detailed results.

\section{Experiments}

\subsection{Experimental Setup}

\paragraph{Backbones} We evaluate our method on three LLMs: LLaMA-3.1-8B-Instruct \citep{dubey2024llama}, Qwen-2.5-7B-Instruct \citep{yang2024qwen2}, and Gemma-2-9B-it \citep{team2024gemma}.

\paragraph{Dataset}  
We use the ExTES dataset as the gold-standard conversational benchmark for expansion. To create the ESC-Pro dataset, we select the first 100 samples from the processed training set as a high-quality starting point. Leveraging this subset, we apply MCTS to generate the ESC-Pro dataset. Additionally, we extract a fine-tuning subset from ESC-Pro, referred to as ESC-Pro-SFT.

\paragraph{Metrics}
We evaluate ESC performance on the ExTES test set and a strategy test set constructed following \citet{kang2024large}. The evaluation includes four metrics: strategy selection accuracy (macro F1 $\mathcal{Q}$, weighted F1 $\mathcal{Q_W}$), strategy preference bias ($\mathcal{B}$), which measures deviation from ideal strategy distributions, and ROUGE-L (R-L) \citep{lin2004rouge} for assessing response semantics.

\paragraph{Baseline}
We compare CSO with both LoRA-based \citep{hu2022lora} and full-parameter supervised fine-tuning (SFT) models trained on ESC-Pro-SFT, following recent ESC approaches \citep{chen2023soulchat,qiu2023smile,zheng2024self}. We also evaluate decoding strategies such as Direct-Refine, Self-Refine, and in-context example prompting. In addition, we benchmark against strong proprietary models, including GPT-4o, Gemini-2.0, and the Claude series. See Appendix~\ref{app:baseline} for details.

\paragraph{Implementation Details}
All experiments are conducted using PyTorch \citep{paszke2019pytorch} on 8 NVIDIA Tesla A100 GPUs. Further details on hyperparameters and training configurations are provided in Appendix \ref{implementation_details}.

\subsection{Overall Results}

\begin{table}[!t]
\centering
\scriptsize
\setlength{\extrarowheight}{0pt}
\renewcommand{\arraystretch}{1.2}
\resizebox{\linewidth}{!}{
\begin{tabular}{l c c c c c}
\toprule
\multicolumn{2}{c}{\textbf{}} & \multicolumn{1}{c}{$\mathcal{Q}\uparrow$} & \multicolumn{1}{c}{$\mathcal{B}\downarrow$} & \multicolumn{1}{c}{\textbf{$\mathcal{Q_W}\uparrow$}} & \multicolumn{1}{c}{\textbf{R - L$\uparrow$}}\\
\midrule
GPT-4o-mini &	&35.68	&2.65	&42.08	&24.91 \\
GPT-4o &	&27.98	&2.65	&39.55	&24.26 \\
Gemini-2.0-Pro &	&27.00	&1.90	&46.59	&22.71 \\
Claude-3.5-Sonnet &	&20.97	&1.55	&41.00	&19.82 \\
Claude-3.7-Sonnet &	&31.50	&1.18	&48.13	&21.01 \\
\midrule
\multicolumn{2}{l}{\textbf{LLaMA-3.1-8B-Instruct}} & \cellcolor{white}29.79 & \cellcolor{white}1.18 & \cellcolor{white}38.78 & \cellcolor{white}23.48 \\
\midrule
\multicolumn{2}{l}{Direct-Refine} & \cellcolor{mycolor_red}{16.08} & \cellcolor{mycolor_red}{1.60} & \cellcolor{mycolor_red}{18.56} & \cellcolor{mycolor_red}{19.53} \\
\multicolumn{2}{l}{Self-Refine} & \cellcolor{mycolor_red}{17.85} & \cellcolor{mycolor_red}{1.35} & \cellcolor{mycolor_red}{24.72} & \cellcolor{mycolor_red}{19.48} \\
\multicolumn{2}{l}{w/ Example} & \cellcolor{mycolor_red}{8.85} & \cellcolor{mycolor_red}{1.27} & \cellcolor{mycolor_red}{15.34} & \cellcolor{mycolor_red}{18.42} \\
\midrule
\multirow{2}{*}{Full} & SFT & \cellcolor{mycolor_green}{\textcolor{black}{30.28}} & \cellcolor{mycolor_red}{\textcolor{black}{2.65}} & \cellcolor{mycolor_red}{\textcolor{black}{37.33}} & \cellcolor{mycolor_red}{\textcolor{black}{23.77}} \\

 & \textbf{CSO} & \cellcolor{mycolor_green}{\textcolor{black}{\textbf{33.11}}} & \cellcolor{mycolor_green}{\textcolor{black}{\textbf{1.11}}} & \cellcolor{mycolor_green}{\textcolor{black}{\textbf{39.21}}} & \cellcolor{mycolor_green}{\textcolor{black}{\textbf{24.24}}} \\
\midrule
\multirow{2}{*}{LoRA} & SFT & \cellcolor{mycolor_green}{31.25} & \cellcolor{mycolor_red}{2.65} & \cellcolor{mycolor_green}{39.27} & \cellcolor{mycolor_red}{23.30} \\

 & \textbf{CSO} & \cellcolor{mycolor_green}{\textbf{34.51}} & \cellcolor{mycolor_green}{\textbf{1.11}} & \cellcolor{mycolor_green}{\textbf{41.11}} & \cellcolor{mycolor_green}{\textbf{23.89}} \\
\midrule
\midrule
\multicolumn{2}{l}{\textbf{Qwen-2.5-7B-Instruct}}  & 19.84 & 2.47 & 28.12 & 23.52 \\
\midrule
\multicolumn{2}{l}{Direct-Refine} & \cellcolor{mycolor_red}12.70 & \cellcolor{mycolor_green}\textbf{1.20} & \cellcolor{mycolor_red}24.89 & \cellcolor{mycolor_red}22.91 \\

\multicolumn{2}{l}{Self-Refine} & \cellcolor{mycolor_red}{11.77} & \cellcolor{mycolor_green}{1.75} & \cellcolor{mycolor_red}{19.59} & \cellcolor{mycolor_red}{20.53} \\

\multicolumn{2}{l}{w/ Example} & \cellcolor{mycolor_red}{17.33} & \cellcolor{mycolor_green}{1.37} & \cellcolor{mycolor_green}{28.21} & \cellcolor{mycolor_red}{22.51} \\
\midrule
\multirow{2}{*}{Full} & SFT    & \cellcolor{mycolor_green}{21.73} & \cellcolor{mycolor_green}{2.34} & \cellcolor{mycolor_green}{31.24} & \cellcolor{mycolor_green}{23.54} \\

& \textbf{CSO} & \cellcolor{mycolor_green}{\textbf{28.78}} & \cellcolor{mycolor_green}{1.92} & \cellcolor{mycolor_green}{\textbf{34.39}} & \cellcolor{mycolor_green}{\textbf{26.16}} \\

\midrule
\multirow{2}{*}{LoRA} & SFT   & \cellcolor{mycolor_green}{21.54} & \cellcolor{mycolor_green}{2.45} & \cellcolor{mycolor_green}{29.11} & \cellcolor{mycolor_green}{23.72} \\

& \textbf{CSO}  & \cellcolor{mycolor_green}{\textbf{23.16}} & \cellcolor{mycolor_green}{2.09} & \cellcolor{mycolor_green}{\textbf{32.26}} & \cellcolor{mycolor_green}\textbf{{24.17}} \\
\midrule
\midrule
\multicolumn{2}{l}{\textbf{Gemma-2-9b-it}}   & 31.31 & 1.33 & 44.06 & 25.64 \\
\midrule
\multicolumn{2}{l}{Direct-Refine} & \cellcolor{mycolor_red}{7.79} & \cellcolor{mycolor_red}{2.55} & \cellcolor{mycolor_red}{12.86} & \cellcolor{mycolor_red}{21.67} \\
\multicolumn{2}{l}{Self-Refine} & \cellcolor{mycolor_red}{15.95} & \cellcolor{mycolor_red}{2.47} & \cellcolor{mycolor_red}{22.93} & \cellcolor{mycolor_red}{20.63} \\
\multicolumn{2}{l}{w/ Example} & \cellcolor{mycolor_red}{20.12} & \cellcolor{mycolor_red}{2.65} & \cellcolor{mycolor_red}{13.41} & \cellcolor{mycolor_red}{19.64} \\
\midrule
\multirow{2}{*}{Full} & SFT  & \cellcolor{mycolor_green}{32.52} & \cellcolor{mycolor_green}{\textbf{1.29}} & \cellcolor{mycolor_green}{46.45} & \cellcolor{mycolor_red}{25.25} \\
& \textbf{CSO} & \cellcolor{mycolor_green}{\textbf{35.61}} & \cellcolor{mycolor_red}{1.54} & \cellcolor{mycolor_green}{\textbf{47.95}} & \cellcolor{mycolor_green}{\textbf{26.63}} \\
\midrule
\multirow{2}{*}{LoRA} & SFT & \cellcolor{mycolor_green}{31.40} & \cellcolor{mycolor_red}{1.55} & \cellcolor{mycolor_red}{43.90} & \cellcolor{mycolor_green}{25.68} \\
& \textbf{CSO}  & \cellcolor{mycolor_green}{\textbf{35.77}} & \cellcolor{mycolor_green}{\textbf{1.23}} & \cellcolor{mycolor_green}{\textbf{52.34}} & \cellcolor{mycolor_green}{\textbf{26.61}} \\
\bottomrule
\end{tabular}}
\caption{Performance comparison of CSO and baseline methods across LLaMA-3.1-8B-Instruct, Qwen2.5-7B-Instruct, and Gemma-2-9B-it backbones in both LoRA and full fine-tuning settings. $\uparrow$ indicates higher is better, $\downarrow$ indicates lower is better.}
\label{tab:main_results}
\end{table}

\subsubsection{Automatic Evaluation Results}
Table \ref{tab:main_results} demonstrates the performance of \textbf{CSO} and baselines based on LLaMA-3.1-8B-Instruct, Qwen2.5-7B-Instruct and Gemma-2-9B-it.

\textbf{CSO improves strategy accuracy while reducing bias.}  
Table \ref{tab:main_results} shows that CSO significantly enhances strategy accuracy while effectively reducing strategy bias. While SFT without preference optimization slightly improves strategy selection accuracy, it also increases strategy bias, limiting adaptability. In contrast, CSO mitigates bias while boosting accuracy, highlighting the necessity of preference optimization in ESC strategy selection. Our results also highlight the limitations of decoding-based approaches, which fail to achieve stable improvements, often leading to lower strategy accuracy and increased bias. Notably, CSO-equipped open-weight models outperform several leading closed-source models on both accuracy and bias, demonstrating its strong generalization.

\textbf{CSO enhances ESC performance across different models.}  
CSO improves ESC performance across all backbone models. It enhances strategy accuracy and reduces bias in both weaker models (Qwen2.5-7B-Instruct) and stronger models (Gemma-2-9B-it), demonstrating its versatility and robustness. Moreover, we further evaluate CSO on a larger model, Qwen2.5-32B, and observe consistent improvements, confirming the scalability of our approach. Detailed results and analysis are provided in Appendix \ref{app:large_model}.

\textbf{CSO excels in both LoRA and Full fine-tuning settings.}  
CSO consistently outperforms SFT in both LoRA-based fine-tuning and full-parameter fine-tuning. While SFT improvements are more limited in LoRA settings, CSO maintains strong performance even with fewer trainable parameters, making it a practical choice for resource-constrained scenarios.

\subsubsection{Human Evaluation Results}

\begin{table}
    \centering
    \small
    \begin{tabular}{l c c c c}
        \toprule 
        CSO vs. SFT & win & lose & tie & $\kappa$ \\
        \midrule
        Acceptance & \textbf{68.00} & 20.33 & 11.67 & 0.65 \\
        Effectiveness & \textbf{58.33} & 16.00 & 25.67 & 0.55 \\
        Sensitivity & \textbf{60.67} & 21.67 & 17.66 & 0.61 \\
        Satisfaction & \textbf{62.34} & 19.33 & 18.33 & 0.64 \\
        \bottomrule
    \end{tabular}
    \caption{Human evaluation comparing CSO and SFT. Win indicates CSO-generated responses are preferred, while Lose represents cases where SFT responses are rated higher. Tie indicates no preference.}
    \label{tab:human_evaluation_results}
\end{table}

Results in Table \ref{tab:human_evaluation_results} show that CSO consistently outperforms SFT across all human evaluation metrics. Specifically, CSO achieves higher Acceptance (68.00\% win rate), Effectiveness (58.33\%), and Sensitivity (60.67\%), indicating that its responses are more appropriate, impactful, and emotionally attuned. Inter-rater agreement ($\kappa$ scores between 0.55–0.65) indicates a moderate to high level of consistency among evaluators. These findings further validate that preference optimization enhances ESC performance, making responses more empathetic and aligned with user needs. More details of human evaluation can be found in \ref{app:human_eval}.

\begin{figure*}
    \centering
    \includegraphics[width=1\linewidth]{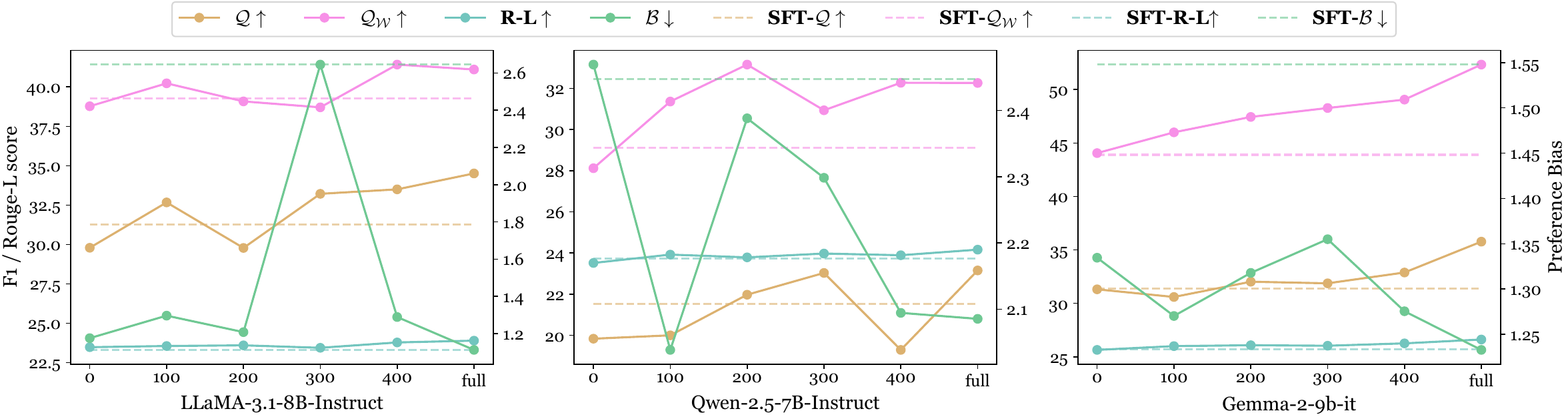}
    \caption{Impact of ESC-Pro data volume on model performance across three backbones. Data size varies with 0 (baseline), 100, 200, 300, 400, and the full set (423 dialogues).}
    \label{fig:data_volume}
\end{figure*}


\subsection{Ablation Study}

\begin{table}
\centering
\scriptsize
\setlength{\extrarowheight}{0pt}
\renewcommand{\arraystretch}{1.2}
\resizebox{\linewidth}{!}{
\begin{tabular}{l  c  c  c  c}
\toprule
\textbf{} & \multicolumn{1}{c}{$\mathcal{Q}\uparrow$} & \multicolumn{1}{c}{$\mathcal{B}\downarrow$} & \multicolumn{1}{c}{\textbf{$\mathcal{Q_W}\uparrow$}} & \multicolumn{1}{c}{\textbf{R - L$\uparrow$}}\\
\midrule
\textbf{LLaMA-3.1-8B-Instruct} & 29.79 & 1.18 & 38.77 & 23.48 \\
\midrule
SFT & \cellcolor{mycolor_green}{31.25} & \cellcolor{mycolor_red}{2.65} & \cellcolor{mycolor_green}{39.27} & \cellcolor{mycolor_red}{23.30} \\
CSO & \cellcolor{mycolor_green}{\textbf{34.51}} & \cellcolor{mycolor_green}{\textbf{1.11}} & \cellcolor{mycolor_green}{\textbf{41.11}} & \cellcolor{mycolor_green}{\textbf{23.89}} \\
CSO - Random & \cellcolor{mycolor_green}{31.79} & \cellcolor{mycolor_red}{2.65} & \cellcolor{mycolor_green}{39.24} & \cellcolor{mycolor_green}{23.65} \\
\midrule
\midrule
\textbf{Qwen-2.5-7B-Instruct} & 19.83 & 2.47 & 28.12 & 23.52 \\
\midrule
SFT & \cellcolor{mycolor_green}{21.54} & \cellcolor{mycolor_green}{2.45} & \cellcolor{mycolor_green}{29.11} & \cellcolor{mycolor_green}{23.72} \\
CSO & \cellcolor{mycolor_green}{\textbf{23.16}} & \cellcolor{mycolor_green}{\textbf{2.09}} & \cellcolor{mycolor_green}{32.26} & \cellcolor{mycolor_green}{\textbf{24.17}} \\
CSO - Random & \cellcolor{mycolor_green}{22.89} & \cellcolor{mycolor_green}{2.19} & \cellcolor{mycolor_green}{\textbf{32.97}} & \cellcolor{mycolor_green}{23.90} \\
\midrule
\midrule
\textbf{Gemma-2-9b-it} & 31.31 & 1.33 & 44.06 & 25.64 \\
\midrule
SFT & \cellcolor{mycolor_green}{31.40} & \cellcolor{mycolor_red}{1.55} & \cellcolor{mycolor_red}{43.90} & \cellcolor{mycolor_red}{25.68} \\
CSO & \cellcolor{mycolor_green}{\textbf{35.77}} & \cellcolor{mycolor_green}{1.23} & \cellcolor{mycolor_green}{\textbf{52.34}} & \cellcolor{mycolor_green}{\textbf{26.61}} \\
CSO - Random& \cellcolor{mycolor_red}{29.86} & \cellcolor{mycolor_green}{\textbf{1.22}} & \cellcolor{mycolor_green}{44.58} & \cellcolor{mycolor_green}{25.75} \\
\bottomrule
\end{tabular}}
\caption{Ablation study results comparing SFT, CSO, and CSO-Random across LLaMA-3.1-8B-Instruct, Qwen2.5-7B-Instruct, and Gemma-2-9B-it.}
\label{tab:ablation_results}
\end{table}

To assess the effectiveness of our approach, we conduct an ablation study using a randomized dataset, ESC-Pro-Random. In this variant, for each Supporter turn, we randomly select a non-preferred strategy instead of using low-scoring strategies from the search process. A non-preferred reply is then generated based on this strategy, forming a preference pair with the original response.

We fine-tune the model using LoRA with the same hyperparameters as ESC-Pro and compare ESC-Pro-Random with both ESC-Pro and standard SFT. As shown in Table \ref{tab:ablation_results}, ESC-Pro-Random performs slightly better than SFT but remains consistently inferior to ESC-Pro across all backbone models. This confirms that our method of leveraging low-scoring nodes searched and verified by MCTS as non-preferred data is both effective and meaningful, reinforcing the importance of structured preference learning in ESC.

\subsection{Data Volume Analysis}

We analyze the impact of data volume by varying the number of ESC-Pro dialogues used for fine-tuning. We experiment with 0, 100, 200, 300, 400, and the full set (423 dialogues), applying LoRA fine-tuning under consistent experimental settings. Figure \ref{fig:data_volume} presents the results, with SFT performance indicated by dashed lines for reference.

The results show a clear upward trend: as data volume increases, performance improves. Notably, with only 200–300 dialogues, CSO outperforms standard SFT, demonstrating the efficiency and scalability of preference data. While further improvements are expected with larger datasets, we limit our seed data to 100 dialogues due to computational constraints. Expanding to larger datasets remains an important direction for future research.

\subsection{Analysis of Different Preference Optimization Algorithms}

To further validate the efficacy of CSO, we examine the impact of integrating alternative preference learning methods into our framework. We replace the default DPO with various existing alternatives, including IPO \citep{azar2024general}, KTO \citep{ethayarajh2024kto}, SimPO \citep{meng2024simpo} and ORPO \citep{hong2024orpo}, and evaluate their performance. The results consistently show that all preference optimization variants outperform standard SFT, reinforcing the advantages of preference-driven learning in ESC. For detailed results and discussion, please refer to Appendix \ref{app:other_pre}.

\section{Conclusion}
In this work, we address low strategy selection accuracy and preference bias challenges in ESC. We propose a novel two-stage framework that optimizes strategy selection preferences at each dialogue turn. We first expand existing ESC datasets with Monte Carlo Tree Search, constructing a conversation tree where different strategy-response pairs are evaluated to generate ESC-Pro, a refined preference dataset. By training LLMs on ESC-Pro, Chain-of-Strategy Optimization (CSO) improves strategy accuracy, reduces bias, and enhances adaptability to user emotions. Extensive experiments on LLaMA-3.1-8B-Instruct, Gemma-2-9B-it, and Qwen2.5-7B-Instruct demonstrate that CSO significantly outperforms standard SFT and decoding-based methods, validating the efficacy of turn-level preference modeling in improving ESC quality.

\section*{Limitations}
Despite the effectiveness of CSO in improving strategy selection and preference alignment, our study has several limitations: Due to limited computational resources, our experiments were conducted on mid-scale LLMs (7B–32B parameters). While these models are representative, larger-scale models (e.g., 70B+) could further enhance performance and provide deeper insights into CSO’s scalability. Our ESC-Pro dataset was generated using a seed set of 100 dialogues, expanded through MCTS-based search. While the results demonstrate significant performance gains, a larger seed dataset or alternative expansion strategies (e.g., human-in-the-loop validation) could further enhance the diversity and quality of preference data. Future work should also place more emphasis on personalization \citep{liu2025survey,qiu2025measuring,qiu2025latent} and safety \citep{zhao2025beware} to ensure that CSO-driven dialogue systems are not only effective but also aligned with user-specific needs and robust against potential risks.

\section*{Ethical Considerations}
Our work is intended solely for research purposes and aims to improve the effectiveness of Emotional Support Conversations (ESC) in Large Language Models (LLMs). While CSO enhances strategy selection and adaptability, it is important to recognize the ethical implications of deploying LLMs in emotionally sensitive interactions.

This study is conducted as a technical exploration and is not intended for direct deployment in real-world mental health or counseling applications. The models used in our experiments are not designed to replace professional human support and should not be used as a substitute for licensed therapy or crisis intervention.

While CSO mitigates strategy bias, LLMs can still exhibit undesirable biases inherited from training data. We take precautions by evaluating toxicity levels and ensuring alignment with supportive strategies, but further human oversight and ethical review are necessary before implementation.

Our study does not involve real user data and strictly utilizes publicly available benchmarks. We encourage future research to adhere to ethical AI principles, including transparency, fairness, and accountability, to prevent potential misuse in emotionally sensitive applications.

\section*{Acknowledgments}
We thank the anonymous reviewers for their comments and suggestions. This work was supported by the New Generation Artificial Intelligence-National Science and Technology Major Project 2023ZD0121100, the National Natural Science Foundation of China (NSFC) via grant 62441614 and 62176078, the Fundamental Research Funds for the Central Universities, and the Singapore Ministry of Education (MOE) Academic Research Fund (AcRF) Tier 1 grant (No. MSS24C004).

\bibliography{custom}


\appendix

\section{LLM Definition and Prompt}
\label{sec:llm_def}

We employ four types of LLMs to guide the dialogue strategy generation and evaluation during the MCTS process.

\paragraph{Seeker LLM}  
The Seeker LLM plays the role of a visitor in an empathic dialogue. Based on the dialogue history up to the last Supporter turn, the Seeker generates a response or outputs an end-of-dialogue token when the conversation should be concluded. The Seeker’s action is defined as:
\[
a_{\text{seeker}} := f_{\text{seeker}}(H_{\text{supporter}})
\]
where \( H_{\text{supporter}} \) represents the conversation history up to the last Supporter turn, denoted as \( \{R_{\text{seeker}}^{(0)}, R_{\text{supporter}}^{(0)}, R_{\text{seeker}}^{(1)}, R_{\text{supporter}}^{(1)}, \dots, \\ R_{\text{seeker}}^{(i)}, R_{\text{supporter}}^{(i)}\} \). The function \( f_{\text{seeker}} \) generates the Seeker’s response or an end-of-conversation token when the Seeker decides to end the dialogue.

\begin{figure}
    \centering
    \includegraphics[width=1\linewidth]{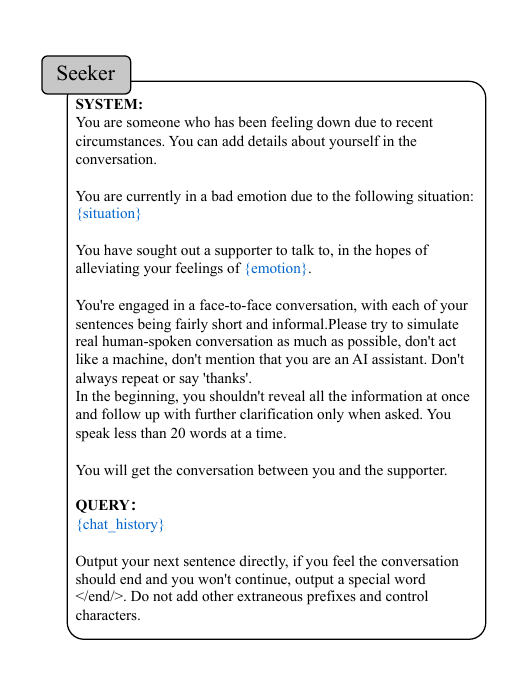}
    \caption{Prompt of seeker LLM.}
    \label{fig:prompt_seeker}
\end{figure}

We use GPT-4o-mini as the Seeker LLM and employ the  prompt shown in Figure \ref{fig:prompt_seeker}.

\paragraph{Supporter LLM}  
The Supporter LLM responds to the Seeker’s turn, offering supportive or complementary dialogue. Based on the Seeker’s last statement and a predefined response strategy, the Supporter generates a reply. The Supporter’s action is defined as:
\[
a_{\text{supporter}} := f_{\text{supporter}}(H_{\text{seeker}}, \text{Strategy})
\]
where \( H_{\text{seeker}} \) represents the conversation history up to the last Seeker turn, denoted as \( \{R_{\text{seeker}}^{(0)}, R_{\text{supporter}}^{(0)}, R_{\text{seeker}}^{(1)}, R_{\text{supporter}}^{(1)}, \dots, \\ R_{\text{supporter}}^{(i-1)}, R_{\text{seeker}}^{(i)}\} \), and \( \text{Strategy} \) refers to the chosen response strategy. The function \( f_{\text{supporter}} \) generates the Supporter’s reply based on these inputs.

\begin{figure}
    \centering
    \includegraphics[width=1\linewidth]{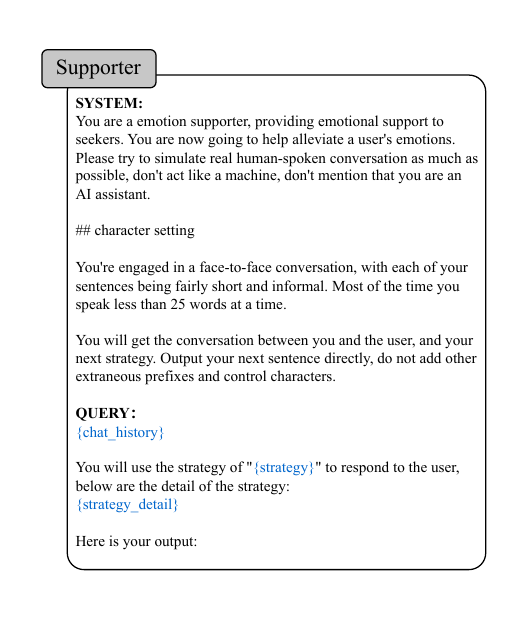}
    \caption{Prompt of supporter LLM.}
    \label{fig:prompt_supporter}
\end{figure}

We use GPT-4o-mini as the Supporter LLM and employ the  prompt shown in Figure \ref{fig:prompt_supporter}.

\paragraph{Strategy LLM}  
The Strategy LLM evaluates available strategies for the Supporter at each node, scoring them on a scale from 1 to 10 based on the Seeker’s last statement. The score reflects the appropriateness and effectiveness of each strategy in the current context. The Strategy’s action is defined as:
\[
a_{\text{strategy}} := f_{\text{strategy}}(H_{\text{seeker}})
\]
where \( H_{\text{seeker}} \) is the conversation history up to the last Seeker turn, as defined above. The output \( a_{\text{strategy}} \) is a score between 1 and 10 for each available strategy, representing its effectiveness in the current context. 

After scoring, the strategy scores are normalized using the softmax function to ensure they form a valid probability distribution across the strategies at the children of the same node. This normalization allows the scores to be used as the \( P \)-value in the PUCB formula. 

\begin{figure}
    \centering
    \includegraphics[width=1\linewidth]{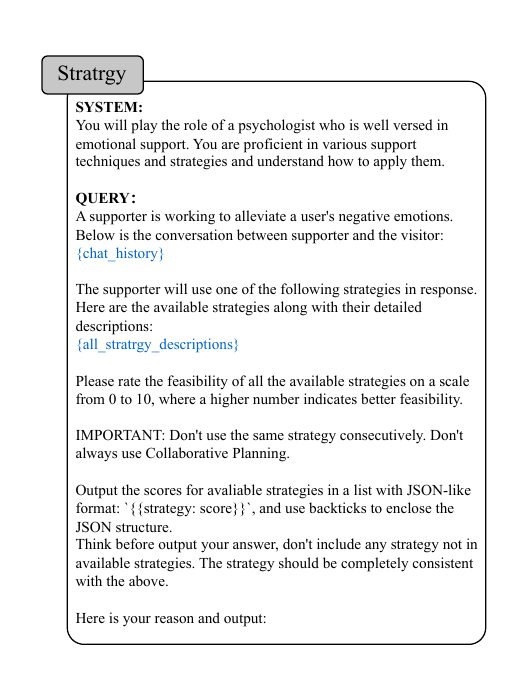}
    \caption{Prompt of strategy LLM.}
    \label{fig:prompt_strategy}
\end{figure}

We use GPT-4o-mini as the Strategy LLM and employ the prompt shown in Figure \ref{fig:prompt_strategy}.

\paragraph{Reward LLM}  
The Reward LLM evaluates the quality of the conversation based on four metrics: Empathy (\( E \)), Information (\( I \)), Humanoid (\( H \)), and Strategies (\( S \)). The action of the Reward LLM is defined as:
\[
a_{\text{reward}} := f_{\text{reward}}(H_{\text{supporter}})
\]
where \( H_{\text{supporter}} \) is the conversation history up to the last Supporter turn, as defined above. The function \( f_{\text{reward}} \) returns the scores for each of the four metrics, which are represented as a tuple:
\[
f_{\text{reward}}(H^{k}_{\text{supporter}}) = (E^{k}, I^{k}, H^{k}, S^{k})
\]

We use GPT-4o-mini as the Reward LLM and employ the  prompt shown in Figure \ref{fig:prompt_reward}. Detailed evaluation criteria are in the Appendix \ref{app:reward_eval}.

\begin{figure}[!t]
    \centering
    \includegraphics[width=1\linewidth]{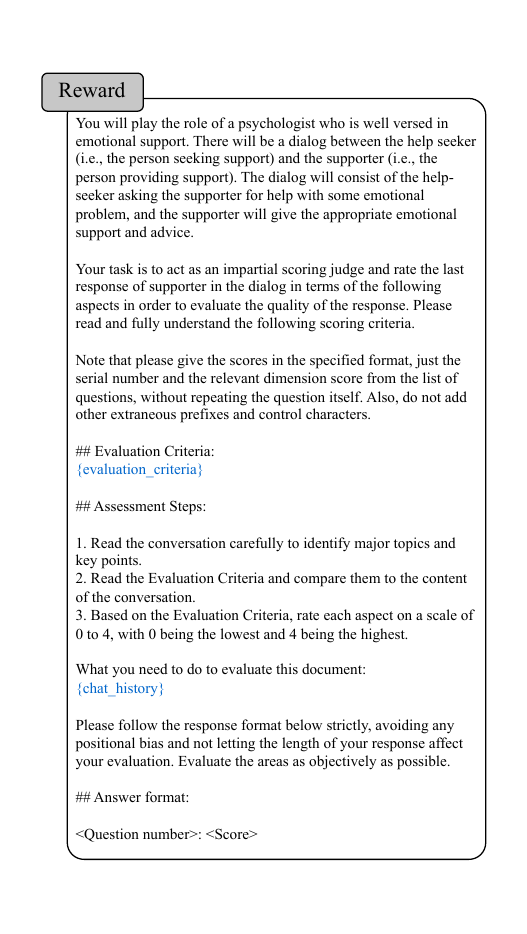}
    \caption{Prompt of reward LLM.}
    \label{fig:prompt_reward}
\end{figure}

\section{Reward Evaluation Criteria}
\label{app:reward_eval}

The explanations of each metric are as follows:

\paragraph{Empathy (E)} Focusing on the comprehension of user emotions and the delineation of the underlying logical framework of user emotions.

\paragraph{Information (I)} Focusing on Evaluating the Reasonableness and Quantity of Recommendations Provided by Emotion Assistants.

\paragraph{Humanoid (H)} Focus on the differences between emotional assistants and humans.

\paragraph{Strategies (S)} Evaluating the Accuracy and Appropriateness of Emotional Support Strategies Used by Assistants.

Evaluation rules are listed in Table \ref{tab:evaluation_criteria}.

\begin{table}
    \centering
    \resizebox{\linewidth}{!}{
    \begin{tabular}{l rrrrr}
        \toprule 
        & ExTES & ESC-Pro(seeker) & ESC-Pro(+) & ESC-Pro(-) & ESC-Pro \\
        \midrule
        Dialogues & 100 & $\backslash$ & $\backslash$ & $\backslash$ & 423 \\
        Utterances & 1,613 & 3,113 & 3,113 & 8,157 & 14,383 \\
        Avg. len. of dialogues & 16.13 & $\backslash$ & $\backslash$ & $\backslash$ & 14.72 \\
        Avg. len. of utterances & 29.03 & 17.33 & 29.42 & 23.22 & 23.29 \\
        \bottomrule
    \end{tabular}}
    \caption{The data statics of our ESC-Pro.}
    \label{tab:data_characteristics}
\end{table}

\section{Dataset Evaluation}

\subsection{Statics of ESC-Pro}
\label{app:data_statics}
As shown in Table \ref{tab:data_characteristics}, we expand 100 seed dialogues from ExTES \citep{zheng2024self} into 423 dialogues, forming our ESC-Pro dataset. The total number of utterances grows from 1,613 to 14,383, with over half (8,157 utterances) classified as non-preference data. This demonstrates that our method not only effectively expands high-quality preference data but also generates a substantial amount of non-preference data, making ESC-Pro well-suited for preference optimization.

The average dialogue length remains consistent between the expanded dataset (14.72 utterances) and the original (16.13 utterances), ensuring that expansion does not degrade data quality. Additionally, the average length of preference utterances (29.42) closely matches that of the seed data (29.03), while non-preference utterances (23.22) are notably shorter. This distinction highlights the effectiveness of our method in capturing meaningful preference differences within ESC interactions.

\begin{table}
    \centering
    \resizebox{\linewidth}{!}{
    \begin{tabular}{l c c c c c}
       \toprule 
        & ExTES & ESC-Pro(+) & ESC-Pro(-) & ESC-Pro & $\kappa$ \\
        \midrule
        \textbf{Acc} & 3.78 & \textbf{3.91} & 3.13 & 3.52 & 0.41 \\
      
        \textbf{Eff} & 3.31 & \textbf{3.45} & 3.19 & 3.27 & 0.51 \\
      
        \textbf{Sen} & 3.86 & \textbf{3.98} & 3.51 & 3.75 & 0.44 \\
        
       \textbf{Sat} & 3.65 & \textbf{3.78} & 3.28 & 3.52 & 0.45 \\
      
        \textbf{Align} & 3.53 & \textbf{3.60} & 3.47 & 3.54 & 0.52 \\
    \bottomrule
    \end{tabular}}
    \caption{Human evaluation results comparing ExTES-Seed, ESC-Pro(+), ESC-Pro(-), and ESC-Pro. (+) and (-) mean the preferred and dispreferred turns, respectively. Higher scores indicate better response quality.}
    \label{tab:quality_evaluation}
\end{table}

\subsection{Data Quality Evaluation}
\label{app:data_quality}
We conduct a human evaluation on 100 responses from the ExTES seed dataset, along with 50 preference responses and 50 non-preference responses from ESC-Pro. We consider five metrics \citep{kang2024large}:
\begin{itemize}[]
    \item Acceptance (Acc): Measures the response’s general acceptability.
    \item Effectiveness (Eff): Assesses whether the response effectively addresses the seeker’s concerns.
    \item Sensitivity (Sen): Evaluates the response’s ability to perceive and respond to emotional cues.
    \item Satisfaction (Sat): Represents the seeker’s overall evaluation, computed as the average of Acc, Eff, and Sen.
    \item Alignment (Align): Assesses whether the response aligns with the intended strategy.
\end{itemize}

As shown in Table \ref{tab:quality_evaluation}, preference data (ESC-Pro+) consistently outperforms non-preference data (ESC-Pro-) across all metrics, with scores slightly exceeding those of the original seed dataset. Notably, the Alignment score for non-preference data is lower, confirming that ineffective strategies contribute to weaker responses. These findings validate ESC-Pro's ability to distinguish and refine high-quality ESC strategies.

\begin{table*}
    \centering
    \resizebox{\linewidth}{!}{
    \begin{tabular}{l c c c c c c}
        \toprule
        & TOXICITY & SEVERE\_TOXICITY & IDENTITY\_ATTACK & INSULT & PROFANITY & THREAT \\
        \midrule
        ExTES(seeker) & 0.0281 & 0.0012 & 0.0041 & 0.0108 & 0.0171 & 0.0088 \\
     
        ExTES(supporter) & \textbf{0.0173} & \textbf{0.0008} & \textbf{0.0027} & \textbf{0.0093} & \textbf{0.0124} & \textbf{0.0070} \\
       
        ExTES & 0.0227 & 0.0010 & 0.0034 & 0.0100 & 0.0148 & 0.0079 \\
        
        ESC - Pro(seeker) & 0.0290 & 0.0013 & 0.0042 & 0.0110 & 0.0178 & 0.0097 \\
       
        ESC - Pro(+) & 0.0192 & \textbf{0.0008} & \textbf{0.0027} & 0.0097 & 0.0130 & 0.0074 \\
      
        ESC - Pro(-) & 0.0223 & 0.0009 & \textbf{0.0027} & 0.0107 & 0.0134 & 0.0074 \\
        
        ESC - Pro & 0.0231 & 0.0010 & 0.0030 & 0.0105 & 0.0143 & 0.0079 \\
    \bottomrule
    \end{tabular}}
    \caption{Toxicity analysis of different results.}
    \label{tab:toxic_analysis}
\end{table*}

\begin{figure}
    \centering
    \includegraphics[width=1\linewidth]{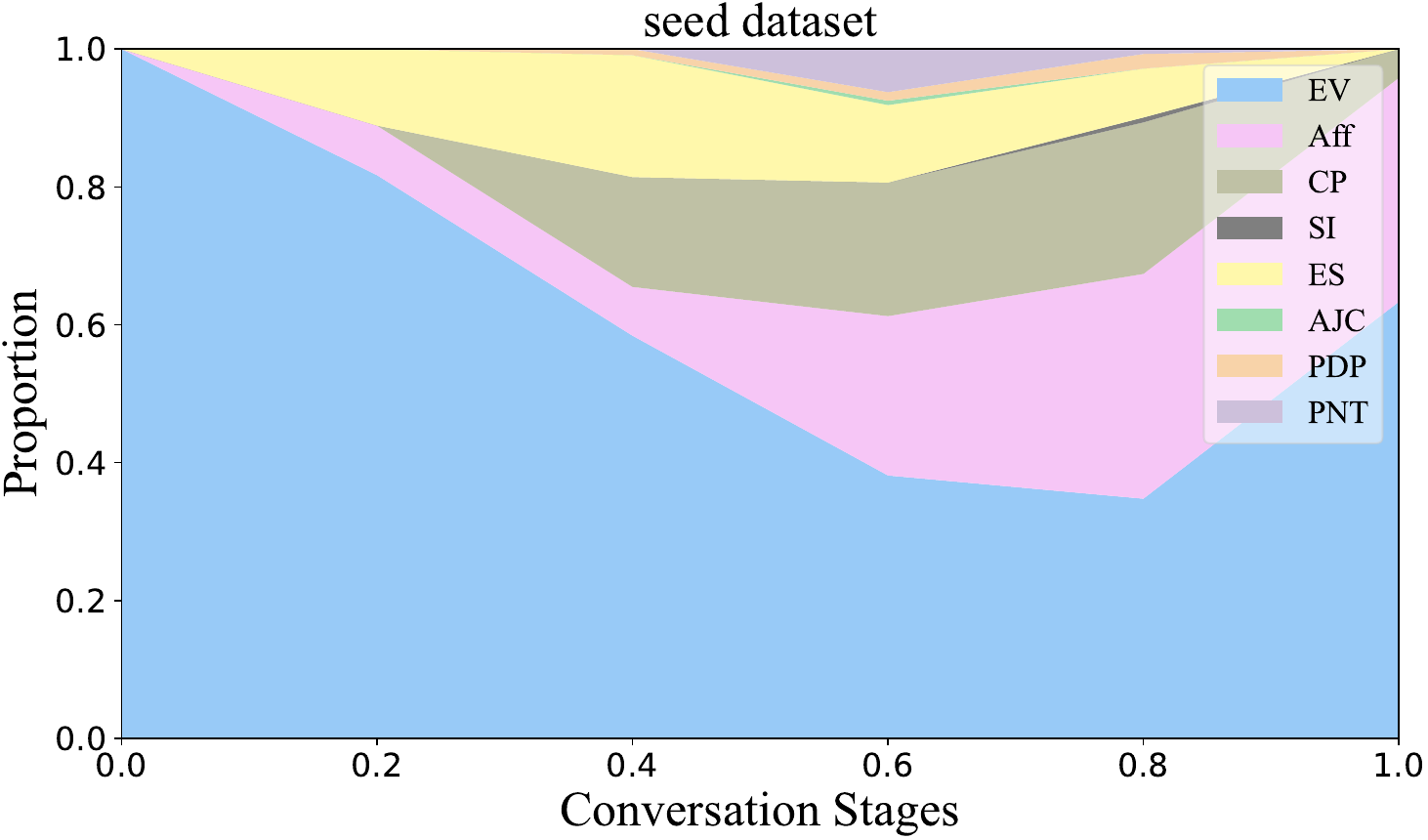}
    \caption{Strategy distribution across conversation stages in the seed dataset.}
    \label{fig:strategy_distribution_seed}
\end{figure}

\subsection{Strategy Analysis}
\label{app:strategy}
The ExTES dataset originally contains 16 distinct support strategies. To reduce the search space and improve computational efficiency, we merge similar strategies into 8 categories (see Appendix \ref{app:strategies_def} for details). We analyze the distribution of these strategies across six conversation stages. Given a dialogue with $N$ utterances, each utterance $k$ (where $k = 1, …, N$) is assigned to stage $i$ using:
\[
i = \left\lfloor \frac{k}{N} \times 6 \right\rfloor \times 0.2
\]
where $i$ ranges from 0 to 1 in increments of 0.2.

As shown in Figure \ref{fig:strategy_distribution}, the ESC-Pro dataset exhibits a dynamic and contextually appropriate strategy distribution. For instance, ``Emotional Validation'' is most prevalent in the early stages, helping to acknowledge and empathize with the seeker, but its usage declines in later stages. In contrast, ``Affirmation'' and ``Avoid Judgment and Criticism'' become more frequent toward the end, offering support and encouragement.

A comparison with the seed dataset (Figure \ref{fig:strategy_distribution_seed}) reveals that ESC-Pro employs a more diverse and balanced strategy distribution. This indicates that our method effectively models varied dialogue strategies, enriching the dataset beyond what was originally present in ExTES.

\subsection{Toxicity Analysis}
\label{app:toxicity}

We assess the toxicity levels of ESC-Pro using the Perspective API, a widely used tool for detecting harmful content. Table \ref{tab:toxic_analysis} summarizes the results across six toxicity attributes.

Our analysis shows that ESC-Pro maintains a similar toxicity profile to ExTES-seed, ensuring its suitability for preference optimization. Specifically: Preference responses (ESC-Pro+) exhibit lower toxicity than non-preference responses (ESC-Pro-), indicating that high-quality strategy selection leads to less harmful outputs. Seeker utterances in both ESC-Pro and ExTES-seed show relatively higher toxicity, which aligns with expectations, as they often reflect negative emotions or distress typical in emotional support dialogues. Supporter responses in ESC-Pro remain within a reasonable toxicity range, demonstrating that preference-based expansion does not introduce significant risks.

Overall, these findings confirm that ESC-Pro effectively balances strategy expansion while maintaining low toxicity, making it well-suited for safe and supportive ESC generation.

\section{Definitions of Strategies}
\label{app:strategies_def}

The reduction from 16 to 8 strategies is a deliberate design decision based on both empirical observations and theoretical coherence. Many of the original strategies in \citet{zheng2024self} have significant semantic overlap and are frequently confused by annotators in preliminary labeling exercises. Detailed categories are shown in Table \ref{tab:strategy_map}.

For example: Reflective Statements, Clarification, Normalize Experiences, and Emotional Validation all aim to validate or mirror the user's feelings. These are consistently grouped together under the broader category of Emotional Validation by human annotators. Similarly, Suggest Options, Collaborative Planning, Stress Management, and Promote Self-Care Practices are all oriented toward joint problem-solving, and are thus unified under Collaborative Planning.

We make these groupings with the dual goal of: (1) Improving label consistency and inter-annotator agreement (IAA), and (2) Focusing the learning signal on distinct and meaningful strategic differences rather than fine-grained variations that models (and even humans) struggle to reliably differentiate.

\paragraph{Emotional Validation (EV)} Acknowledge and validate the User’s emotions without judgment.

\paragraph{Affirmation (Aff)} Provide positive reinforcement and encouragement to uplift the User’s spirits.

\paragraph{Collaborative Planning (CP)} Work together with the User to develop an action plan.

\paragraph{Empathetic Statements (ES)} Express understanding and empathy towards the User’s experiences.

\paragraph{Avoid Judgment and Criticism (AJC)} It’s important to create a non-judgmental and safe space for the User to express their emotions without fear of criticism. Refrain from passing judgment or being overly critical of their experiences or choices.

\paragraph{Provide Different Perspectives (PDP)} Offer alternative ways of looking at the situation to help the User gain new insights.

\paragraph{Reframe Negative Thoughts (RNT)} Help the User reframe negative thoughts into more positive or realistic ones.

\paragraph{Share Information (SI)} Provide educational or factual information about emotions, coping mechanisms, or self-care practices.

\section{Baseline}
\label{app:baseline}

\paragraph{Direct Refine} Direct Refine is a simple and direct self-optimization method, where the model directly refines its output based on the original response. This approach aims to improve the quality of the generated text by making adjustments without additional external input.

\paragraph{Self-Refine} Based on the approach by \citet{madaan2024self}, we employ a two-step process. First, the model is required to reflect on its original output and generate feedback. Then, it uses this feedback to generate an optimized response. This method encourages self-correction and refinement based on the model's own reflections.

\paragraph{w/ Example} In this baseline, we randomly select a sample from the ExTES training set and incorporate it into the prompt. This example serves as a reference for the model, providing additional context to guide its generation process and improve response quality.

\section{Implementation Details}
\label{implementation_details}

\subsection{Dataset Construction Details}

In constructing our ESC-Pro dataset, we set the hyperparameter \( c \) in the PUCB formula used in the MCTS process (Eq.\ref{eq:pucb}) to 1, in order to balance exploration and exploitation. Additionally, we set the hyperparameter \( \alpha \) in the reward calculation formula (Eq.\ref{eq:reward_calc}) to 7, and the hyperparameter \( b \) to -3, which directs the search towards strategies with a higher degree of rationality and encourages the algorithm to prioritize nodes with scores greater than 3. When extracting data from the dialogue tree (Eq.\ref{eq:path}, Eq.\ref{eq:dataset}), we set the score threshold \( \theta \) to 0.5, ensuring that the scores of the preference data are at least greater than 3.5, thereby maintaining the quality of the dataset. 

When processing the seed data from ExTES, we filtered out entries that did not provide a strategy, contained incorrect strategies, or included the ``Others'' strategy. The remaining 15 strategies (excluding ``Others'') used in the ExTES dataset were mapped to the 8 strategies used in our dataset according to a set of predefined rules, which can be found in Table \ref{tab:strategy_map}.

\begin{table}
    \centering
    \scriptsize
    \setlength{\extrarowheight}{0pt}
    \renewcommand{\arraystretch}{1.2}
    \resizebox{\linewidth}{!}{
        \begin{tabular}{c c c}
            \toprule
            Reflective Statements & \( \rightarrow \) & Emotional Validation\\
            Clarification & \( \rightarrow \) & Emotional Validation\\
            Emotional Validation & \( \rightarrow \) & Emotional Validation\\
            Normalize Experiences & \( \rightarrow \) & Emotional Validation\\
            \midrule
            Affirmation & \( \rightarrow \) & Affirmation\\
            Offer Hope & \( \rightarrow \) & Affirmation\\
            \midrule
            Suggest Options & \( \rightarrow \) & Collaborative Planning\\
            Collaborative Planning & \( \rightarrow \) & Collaborative Planning\\
            Stress Management & \( \rightarrow \) & Collaborative Planning\\
            Promote Self-Care Practices & \( \rightarrow \) & Collaborative Planning\\
            \midrule
            Empathetic Statements & \( \rightarrow \) & Empathetic Statements\\
            \midrule
            Avoid Judgment and Criticism & \( \rightarrow \) & Avoid Judgment and Criticism\\
            \midrule
            Provide Different Perspectives & \( \rightarrow \) & Provide Different Perspectives\\
            \midrule
            Reframe Negative Thoughts & \( \rightarrow \) & Reframe Negative Thoughts\\
            \midrule
            Share Information & \( \rightarrow \) & Share Information\\
            \bottomrule
        \end{tabular}
    }
    \caption{Strategy Mapping Rules}
    \label{tab:strategy_map}
\end{table}

\subsection{Experimental Details}

Our experiments are implemented with PyTorch \citep{paszke2019pytorch} on 8 NVIDIA Tesla A100 using DeepSpeed \citep{rasley2020deepspeed} repository with ZeRo-2 optimization. We performed both full-parameter and LoRA fine-tuning on LLaMA-3.1-8B-Instruct \citep{dubey2024llama}, Qwen-2.5-7B-Instruct \citep{yang2024qwen2} and Gemma-2-9B-it \citep{team2024gemma}. For all experiments, we set maximum target length of 512 tokens across all backbones. LoRA fine-tuning is conducted with an alpha of 8, and a dropout rate of 0, targeting all modules. All backbones are trained using their respective official chat templates.

More detailed settings across different backbones are listed in Table \ref{tab:hyper} and Table \ref{tab:hyper_beta}.

\section{Additional Experimental Results}

\subsection{Analysis of Different Preference Optimization Algorithms}
\label{app:other_pre}

\begin{table}
    \centering
    \scriptsize
    \setlength{\extrarowheight}{0pt}
    \renewcommand{\arraystretch}{1.2}
    \resizebox{\linewidth}{!}{
        \begin{tabular}{p{0.5cm}p{2cm} c c c c}
            \toprule
            \multicolumn{2}{c}{\textbf{}} & \multicolumn{1}{c}{$\mathcal{Q}\uparrow$} & \multicolumn{1}{c}{$\mathcal{B}\downarrow$} & \multicolumn{1}{c}{\textbf{$\mathcal{Q_W}\uparrow$}} & \multicolumn{1}{c}{\textbf{R - L$\uparrow$}}\\
            \midrule
            \multicolumn{2}{l}{\textbf{LLaMA-3.1-8B-Instruct}} & \cellcolor{white}29.79 & \cellcolor{white}1.18 & \cellcolor{white}38.78 & \cellcolor{white}23.48 \\
            \midrule
            \multirow{6}{*}{FuLL} 
             & SFT   & \cellcolor{mycolor_green}{\textcolor{black}{30.28}} & \cellcolor{mycolor_red}{\textcolor{black}{2.65}} & \cellcolor{mycolor_red}{\textcolor{black}{37.33}} & \cellcolor{mycolor_red}{\textcolor{black}{23.77}} \\
             & \textbf{CSO-DPO}  & \cellcolor{mycolor_green}{\textcolor{black}{33.11}} & \cellcolor{mycolor_green}{\textcolor{black}{1.11}} & \cellcolor{mycolor_green}{\textcolor{black}{\textbf{39.21}}} & \cellcolor{mycolor_green}{\textcolor{black}{24.24}} \\
             & \textbf{CSO-SimPO} & \cellcolor{mycolor_green}{\textcolor{black}{29.12}} & \cellcolor{mycolor_red}{\textcolor{black}{1.53}} & \cellcolor{mycolor_red}{\textcolor{black}{36.27}} & \cellcolor{mycolor_green}{\textcolor{black}{23.59}} \\
             & \textbf{CSO-IPO} & \cellcolor{mycolor_green}{\textcolor{black}{\textbf{35.48}}} & \cellcolor{mycolor_green}{\textcolor{black}{1.04}} & \cellcolor{mycolor_red}{\textcolor{black}{37.74}} & \cellcolor{mycolor_green}{\textcolor{black}{24.19}} \\
             & \textbf{CSO-KTO}& \cellcolor{mycolor_green}{\textcolor{black}{32.60}} & \cellcolor{mycolor_green}{\textcolor{black}{\textbf{0.88}}} & \cellcolor{mycolor_red}{\textcolor{black}{36.63}} & \cellcolor{mycolor_green}{\textcolor{black}{25.84}} \\
             & \textbf{CSO-ORPO}  & \cellcolor{mycolor_green}{\textcolor{black}{30.46}} & \cellcolor{mycolor_green}{\textcolor{black}{1.14}} & \cellcolor{mycolor_red}{\textcolor{black}{33.24}} & \cellcolor{mycolor_green}{\textcolor{black}{\textbf{26.06}}} \\
            \midrule
            \multirow{6}{*}{LoRA} 
             & SFT  & \cellcolor{mycolor_green}{31.25} & \cellcolor{mycolor_red}{2.65} & \cellcolor{mycolor_green}{39.27} & \cellcolor{mycolor_red}{23.30} \\
             & \textbf{CSO-DPO} & \cellcolor{mycolor_green}{\textbf{34.51}} & \cellcolor{mycolor_green}{1.11} & \cellcolor{mycolor_green}{\textbf{41.11}} & \cellcolor{mycolor_green}{\textbf{23.89}} \\
             & \textbf{CSO-SimPO} & \cellcolor{mycolor_green}{33.43} & \cellcolor{mycolor_green}{\textbf{1.04}} & \cellcolor{mycolor_green}{40.55} & \cellcolor{mycolor_red}{23.41} \\
             & \textbf{CSO-IPO}   & \cellcolor{mycolor_green}{33.00} & \cellcolor{mycolor_green}{1.13} & \cellcolor{mycolor_green}{39.40} & \cellcolor{mycolor_green}{23.55} \\
             & \textbf{CSO-KTO} & \cellcolor{mycolor_green}{32.80} & \cellcolor{mycolor_green}{\textbf{1.04}} & \cellcolor{mycolor_red}{38.15} & \cellcolor{mycolor_green}{23.70} \\
             & \textbf{CSO-ORPO}  & \cellcolor{mycolor_green}{31.50} & \cellcolor{mycolor_green}{1.17} & \cellcolor{mycolor_green}{39.04} & \cellcolor{mycolor_green}{23.71} \\
            \bottomrule
        \end{tabular}
    }
    \caption{The Results of different preference optimization algorithms on LLaMA-3.1-8B-Instruct.}
    \label{tab:other_method_llama}
\end{table}

\begin{table}[!t]
    \centering
    \scriptsize
    \setlength{\extrarowheight}{0pt}
    \renewcommand{\arraystretch}{1.2}
    \resizebox{\linewidth}{!}{
        \begin{tabular}{p{0.5cm}p{2cm} c c c c}
            \toprule
            \multicolumn{2}{c}{\textbf{}} & \multicolumn{1}{c}{$\mathcal{Q}\uparrow$} & \multicolumn{1}{c}{$\mathcal{B}\downarrow$} & \multicolumn{1}{c}{\textbf{$\mathcal{Q_W}\uparrow$}} & \multicolumn{1}{c}{\textbf{R - L$\uparrow$}}\\
            \midrule
            \multicolumn{2}{l}{\textbf{Qwen-2.5-7B-Instruct}} & 19.84 & 2.47 & 28.12 & 23.52 \\
            \midrule
            \multirow{6}{*}{FuLL} 
            & SFT & \cellcolor{mycolor_green}{21.73} & \cellcolor{mycolor_green}{2.34} & \cellcolor{mycolor_green}{31.24} & \cellcolor{mycolor_green}{23.54} \\
            & \textbf{CSO-DPO} & \cellcolor{mycolor_green}{\textbf{28.78}} & \cellcolor{mycolor_green}{1.92} & \cellcolor{mycolor_green}{34.39} & \cellcolor{mycolor_green}{26.16} \\
            & \textbf{CSO-SimPO} & \cellcolor{mycolor_green}{23.51} & \cellcolor{mycolor_green}{2.01} & \cellcolor{mycolor_green}{34.84} & \cellcolor{mycolor_red}{21.54} \\
            & \textbf{CSO-IPO} & \cellcolor{mycolor_green}{25.10} & \cellcolor{mycolor_green}{2.15} & \cellcolor{mycolor_green}{36.54} & \cellcolor{mycolor_green}{25.16} \\
            & \textbf{CSO-KTO} & \cellcolor{mycolor_green}{25.56} & \cellcolor{mycolor_green}{1.56} & \cellcolor{mycolor_green}{\textbf{38.44}} & \cellcolor{mycolor_green}{\textbf{26.65}} \\
            & \textbf{CSO-ORPO} & \cellcolor{mycolor_green}{22.20} & \cellcolor{mycolor_green}{\textbf{1.24}} & \cellcolor{mycolor_green}{35.91} & \cellcolor{mycolor_green}{24.58} \\
            \midrule
            \multirow{6}{*}{LoRA} 
            & SFT & \cellcolor{mycolor_green}{21.54} & \cellcolor{mycolor_green}{2.45} & \cellcolor{mycolor_green}{29.11} & \cellcolor{mycolor_green}{23.72} \\
            & \textbf{CSO-DPO} & \cellcolor{mycolor_green}{23.16} & \cellcolor{mycolor_green}{2.09} & \cellcolor{mycolor_green}{\textbf{32.26}} & \cellcolor{mycolor_green}{\textbf{24.17}} \\
            & \textbf{CSO-SimPO} & \cellcolor{mycolor_green}{\textbf{25.91}} & \cellcolor{mycolor_green}{\textbf{2.02}} & \cellcolor{mycolor_green}{30.45} & \cellcolor{mycolor_red}{23.32} \\
            & \textbf{CSO-IPO} & \cellcolor{mycolor_green}{22.71} & \cellcolor{mycolor_green}{2.21} & \cellcolor{mycolor_green}{28.75} & \cellcolor{mycolor_red}{23.49} \\
            & \textbf{CSO-KTO} & \cellcolor{mycolor_green}{22.91} & \cellcolor{mycolor_green}{2.17} & \cellcolor{mycolor_green}{30.82} & \cellcolor{mycolor_green}{23.61} \\
            & \textbf{CSO-ORPO} & \cellcolor{mycolor_green}{22.49} & \cellcolor{mycolor_green}{2.12} & \cellcolor{mycolor_green}{28.42} & \cellcolor{mycolor_red}{23.35} \\
            \bottomrule
        \end{tabular}
    }
    \caption{The Results of different preference optimization algorithms on Qwen-2.5-7B-Instruct.}
    \label{tab:other_method_qwen}
\end{table}

\begin{table}[!t]
    \centering
    \scriptsize
    \setlength{\extrarowheight}{0pt}
    \renewcommand{\arraystretch}{1.2}
    \resizebox{\linewidth}{!}{
        \begin{tabular}{p{0.5cm}p{2cm} c c c c}
            \toprule
            \multicolumn{2}{c}{\textbf{}} & \multicolumn{1}{c}{$\mathcal{Q}\uparrow$} & \multicolumn{1}{c}{$\mathcal{B}\downarrow$} & \multicolumn{1}{c}{\textbf{$\mathcal{Q_W}\uparrow$}} & \multicolumn{1}{c}{\textbf{R - L$\uparrow$}}\\
            \midrule
            \multicolumn{2}{l}{\textbf{Gemma-2-9b-it}} & 31.31 & 1.33 & 44.06 & 25.64 \\
            \midrule
            \multirow{6}{*}{Full} 
            & SFT & \cellcolor{mycolor_green}{32.52} & \cellcolor{mycolor_green}{1.29} & \cellcolor{mycolor_green}{46.45} & \cellcolor{mycolor_red}{25.25} \\
            & \textbf{CSO-DPO} & \cellcolor{mycolor_green}{35.61} & \cellcolor{mycolor_red}{1.54} & \cellcolor{mycolor_green}{47.95} & \cellcolor{mycolor_green}{26.63} \\
            & \textbf{CSO-SimPO} & \cellcolor{mycolor_red}{26.67} & \cellcolor{mycolor_red}{2.03} & \cellcolor{mycolor_green}{48.03} & \cellcolor{mycolor_red}{25.60} \\
            & \textbf{CSO-IPO} & \cellcolor{mycolor_green}{32.02} & \cellcolor{mycolor_red}{1.70} & \cellcolor{mycolor_green}{45.29} & \cellcolor{mycolor_green}{25.81} \\
            & \textbf{CSO-KTO} & \cellcolor{mycolor_green}{\textbf{39.73}} & \cellcolor{mycolor_green}{\textbf{0.81}} & \cellcolor{mycolor_green}{\textbf{48.87}} & \cellcolor{mycolor_green}{\textbf{27.84}} \\
            & \textbf{CSO-ORPO} & \cellcolor{mycolor_green}{34.80} & \cellcolor{mycolor_green}{1.24} & \cellcolor{mycolor_green}{48.28} & \cellcolor{mycolor_green}{27.52} \\
            \midrule
            \multirow{6}{*}{LoRA} 
            & SFT & \cellcolor{mycolor_green}{31.40} & \cellcolor{mycolor_red}{1.55} & \cellcolor{mycolor_red}{43.90} & \cellcolor{mycolor_green}{25.68} \\
            & \textbf{CSO-DPO} & \cellcolor{mycolor_green}{35.77} & \cellcolor{mycolor_green}{1.23} & \cellcolor{mycolor_green}{\textbf{52.34}} & \cellcolor{mycolor_green}{\textbf{26.61}} \\
            & \textbf{CSO-SimPO} & \cellcolor{mycolor_green}{34.95} & \cellcolor{mycolor_green}{1.19} & \cellcolor{mycolor_green}{51.62} & \cellcolor{mycolor_green}{26.22} \\
            & \textbf{CSO-IPO} & \cellcolor{mycolor_green}{34.16} & \cellcolor{mycolor_green}{1.29} & \cellcolor{mycolor_green}{51.62} & \cellcolor{mycolor_green}{26.20} \\
            & \textbf{CSO-KTO} & \cellcolor{mycolor_green}{\textbf{35.89}} & \cellcolor{mycolor_green}{\textbf{0.99}} & \cellcolor{mycolor_green}{48.53} & \cellcolor{mycolor_green}{26.45} \\
            & \textbf{CSO-ORPO} & \cellcolor{mycolor_green}{32.35} & \cellcolor{mycolor_green}{1.30} & \cellcolor{mycolor_green}{48.14} & \cellcolor{mycolor_green}{25.85} \\
            \bottomrule
        \end{tabular}
    }
    \caption{Result of different preference optimization algorithms on Gemma-2-9b-it.}
    \label{tab:other_method_gemma}
\end{table}

In addition to preference training based on DPO, we also investigated the performance of several other preference optimization algorithms, including SimPO \citep{meng2024simpo}, IPO \citep{azar2024general}, KTO \citep{ethayarajh2024kto}, and ORPO \citep{hong2024orpo}. We tested these algorithms in both LoRA and full fine-tuning settings, maintaining the training parameters consistent with DPO, with adjustments made only to algorithm-specific parameters. For instance, the hyperparameter $\beta$ was adjusted to suit each algorithm, and algorithm-specific parameters, such as Gemma-2 for SimPO, chosen weight and rejected weight for KTO, were tuned accordingly. A detailed list of the hyperparameters used for each algorithm can be found in the Table \ref{tab:hyper} and Table \ref{tab:hyper_beta}.

For all algorithms except KTO, we used the ESC-Pro dataset as the training set, which contains preference pairs directly derived from the original ESC-Pro data. In the case of KTO, the training set was derived by splitting each preference pair in the ESC-Pro dataset and removing duplicates. The experimental results are shown in Table \ref{tab:other_method_llama}, Table \ref{tab:other_method_qwen} and Table \ref{tab:other_method_gemma}. From the results, it is evident that all tested preference optimization algorithms performed effectively after training on the ESC-Pro dataset, with some methods achieving higher performance than DPO. These results validate the efficacy and versatility of the ESC-Pro dataset for optimizing preference-based dialogue strategies.

\begin{table}
\centering
\scriptsize
\setlength{\extrarowheight}{0pt}
\renewcommand{\arraystretch}{1.2}
\resizebox{\linewidth}{!}{
\begin{tabular}{l  c  c  c  c}
\toprule
\textbf{} & \multicolumn{1}{c}{$\mathcal{Q}\uparrow$} & \multicolumn{1}{c}{$\mathcal{B}\downarrow$} & \multicolumn{1}{c}{\textbf{$\mathcal{Q_W}\uparrow$}} & \multicolumn{1}{c}{\textbf{R - L$\uparrow$}}\\
\midrule
\textbf{Qwen2.5-32B-Instruct} & 37.37	&1.40	&41.97	&24.37 \\
\midrule
Direct-Refine	&\cellcolor{mycolor_red}34.87	&\cellcolor{mycolor_red}1.68	&\cellcolor{mycolor_red}40.69	&\cellcolor{mycolor_red}23.14 \\
Self-Refine	&\cellcolor{mycolor_red}14.63	&\cellcolor{mycolor_green}\textbf{0.98}	&\cellcolor{mycolor_red}26.97	&\cellcolor{mycolor_red}21.74 \\
w/ Example	&\cellcolor{mycolor_red}20.66	&\cellcolor{mycolor_red}2.65	&\cellcolor{mycolor_red}24.40	&\cellcolor{mycolor_red}21.93 \\
SFT-LoRA	&\cellcolor{mycolor_green}37.69	&\cellcolor{mycolor_red}1.60	&\cellcolor{mycolor_green}42.09	&\cellcolor{mycolor_green}24.40 \\
\midrule
CSO-LoRA	&\cellcolor{mycolor_green}\textbf{38.53}	&\cellcolor{mycolor_green}1.29	&\cellcolor{mycolor_green}43.95	&\cellcolor{mycolor_green}\textbf{24.65} \\
\bottomrule
\end{tabular}}
\caption{Performance comparison on Qwen2.5-32B-Instruct using LoRA-based preference optimization.}
\label{tab:qwen32b}
\end{table}

\subsection{Results of Large-Scale Backbone}
\label{app:large_model}

We conduct additional experiments on Qwen2.5-32B-Instruct using LoRA-based preference optimization. As shown in Table \ref{tab:qwen32b}, CSO-LoRA consistently outperforms all baselines across all four metrics, and long-term user satisfaction (R-L). Compared to standard supervised fine-tuning (SFT-LoRA), CSO-LoRA improves strategy quality by +0.84 and reduces strategy bias while further enhancing user-centered metrics. These results confirm that CSO remains effective even at larger scales, demonstrating strong scalability and robustness in enhancing both adaptability and emotional intelligence in LLM-based emotional support.

\begin{table*}
    \centering
    \small
    \resizebox{\linewidth}{!}{
    \begin{tabularx}{\textwidth}{|l|X|X|X|X|X|X|}
        \hline
        \textbf{Criteria} & \textbf{Empathy} & \textbf{Information} & \textbf{Humanoid} & \textbf{Strategies} \\
        \hline
        \textbf{4 points} & The system exhibits a high degree of anthropomorphism, going so far as to console users in a friendly manner and assist them in analyzing the underlying logic of emotions. & There are many suggestions, and all of them are effective. & There is no apparent difference from human friends. & The strategies are numerous, well-tailored to the user's emotional state, and demonstrate high empathy and effectiveness in addressing the user's concerns. \\
        \hline
        \textbf{3 points} & Providing emotional comfort during conversations and assisting users in analyzing the underlying logical framework of their emotions. & There are more than five suggestions, but some of them are ineffective. There are fewer than five suggestions, but all of them are very effective. & 1-2 traces can reveal that the AI assistant is a language model. & More than five strategies are provided, but some lack empathy or relevance. Alternatively, fewer than five strategies are shared, but they are highly empathetic and directly address the user's core emotional needs. \\
        \hline
        \textbf{2 points} & The lack of understanding of user emotions or the absence of mechanisms to analyze user emotions are the main factors. & The suggestions are fewer than five, and some suggestions are effective, while others provide numerous suggestions, but none of them touch the root of the problem. & More than two traces can reveal that the AI assistant is a language model. & Fewer than five strategies are provided, and they are a mix of relevant and irrelevant approaches. Alternatively, a large number of strategies are given, but they fail to address the user's emotional root issues. \\
        \hline
        \textbf{1 point} & The lack of understanding of user emotions and the absence of mechanisms to analyze user emotions are the main factors. & Have suggestions but ineffective, as well as no suggestions. & Structured responses, or responses in the form of ’As a large language model’ or robot-like replies. & Strategies are present but lack empathy or relevance. Some may appear dismissive or insufficiently supportive in the context of the user's concerns. \\
        \hline
        \textbf{0 points} & The disregard for user concerns, the absence of assistance in analyzing user issues, and even the imposition of negative effects on user emotions. & Suggestions were provided, but all of them were ineffective, and some even gave advice that could potentially harm the user. & The dialogue exhibits rigidity and lacks comprehension in terms of internalizing the content. & Strategies are counterproductive, exacerbating the user's distress or dismissing their concerns. Some suggestions may inadvertently harm the user's emotional well-being. \\
        \hline
    \end{tabularx}}
    \caption{Evaluation criteria of reward LLM.}
    \label{tab:evaluation_criteria}
    
\end{table*}

\begin{table}[!t]
    \centering
    \scriptsize
    \setlength{\extrarowheight}{0pt}
    \resizebox{\linewidth}{!}{
        \begin{tabular}{c c c c c}
            \toprule
            & & \multicolumn{1}{c}{\textbf{Epoch}} & \multicolumn{1}{c}{\textbf{Batch Size}} & \multicolumn{1}{c}{\textbf{Learning Rate}} \\
            \midrule
            \multicolumn{2}{l}{\textbf{LLaMA-3.1-8B-Instruct}}\\
            \midrule
            \multirow{2}{*}{Full} & SFT & 1 & 32 & 5.0e-7 \\
            & \textbf{CSO} & 3 & 32 & 5.0e-7 \\
            \midrule
            \multirow{2}{*}{LoRA} & SFT & 3 & 32 & 1.0e-6 \\
            & \textbf{CSO} & 3 & 32 & 1.0e-6 \\
            \midrule
            \midrule
            \multicolumn{2}{l}{\textbf{Qwen-2.5-7B-Instruct}}\\
            \midrule
            \multirow{2}{*}{FULL} & SFT & 1 & 32 & 5.0e-7 \\
            & \textbf{CSO} & 3 & 32 & 6.0e-7 \\
            \midrule
            \multirow{2}{*}{LoRA} & SFT & 3 & 32 & 1.0e-6 \\
            & \textbf{CSO} & 1 & 128 & 5.0e-7 \\
            \midrule
            \midrule
            \multicolumn{2}{l}{\textbf{Gemma-2-9b-it}}\\
            \midrule
            \multirow{2}{*}{Full} & SFT & 1 & 32 & 5.0e-7 \\
            & \textbf{CSO} & 1 & 32 & 5.0e-7 \\
            \midrule
            \multirow{2}{*}{LoRA} & SFT & 3 & 32 & 5.0e-7 \\
            & \textbf{CSO} & 3 & 8 & 6.0e-7 \\
            \bottomrule
        \end{tabular}
    }
    \caption{Overall hyper-parameter settings.}
    \label{tab:hyper}
\end{table}

\begin{table}
    \centering
    \scriptsize
    \setlength{\extrarowheight}{0pt}
    \renewcommand{\arraystretch}{1.2}
    \resizebox{\linewidth}{!}{
        \begin{tabular}{p{0.5cm}p{2cm} c c c c }
            \toprule
            & & \multicolumn{1}{c}{\textbf{beta}} & \multicolumn{1}{c}{\textbf{gemma}} & \multicolumn{1}{c}{\textbf{chosen weight}} & \multicolumn{1}{c}{\textbf{rejected weight}}\\
            \midrule
            \multicolumn{2}{l}{\textbf{LLaMA-3.1-8B-Instruct}} \\
            \midrule
            \multirow{5}{*}{FuLL} 
             & \textbf{DPO}   & 0.7 & - & - & -  \\
             & \textbf{SimPO}   & 3.3 & 2.0 & - & -  \\
             & \textbf{IPO}  & 1.0 & - & - & -  \\
             & \textbf{KTO}  & 0.9 & - & 1.0 & 0.5  \\
             & \textbf{ORPO} & 1.3 & - & - & -  \\
            \midrule
            \multirow{5}{*}{LoRA} 
             & \textbf{DPO} & 0.01 & - & - & -  \\
             & \textbf{SimPO} & 2.0 & 0.5 & - & -  \\
             & \textbf{IPO}  & 0.1 & - & - & -  \\
             & \textbf{KTO} & 0.01 & - & 1.0 & 1.0  \\
             & \textbf{ORPO}  & 1.0 & - & - & -  \\
            \midrule
            \midrule
            \multicolumn{2}{l}{\textbf{Qwen-2.5-7B-Instruct}} \\
            \midrule
            \multirow{5}{*}{FuLL} 
             & \textbf{DPO}   & 0.1 & - & - & -  \\
             & \textbf{SimPO}   & 1.5 & 3.5 & - & -  \\
             & \textbf{IPO}  & 0.5 & - & - & -  \\
             & \textbf{KTO}  & 0.07 & - & 1.0 & 0.5  \\
             & \textbf{ORPO} & 0.7 & - & - & -  \\
            \midrule
            \multirow{5}{*}{LoRA} 
             & \textbf{DPO} & 0.1 & - & - & -  \\
             & \textbf{SimPO} & 3.5 & 3.0 & - & -  \\
             & \textbf{IPO}  & 0.01 & - & - & -  \\
             & \textbf{KTO} & 0.05 & - & 1.0 & 0.5  \\
             & \textbf{ORPO}  & 2.0 & - & - & -  \\
            \midrule
            \midrule
            \multicolumn{2}{l}{\textbf{Gemma-2-9b-it}} \\
            \midrule
            \multirow{5}{*}{FuLL} 
             & \textbf{DPO}   & 0.5 & - & - & -  \\
             & \textbf{SimPO}   & 3.5 & 2.0 & - & -  \\
             & \textbf{IPO}  & 1.0 & - & - & -  \\
             & \textbf{KTO}  & 0.5 & - & 1.0 & 0.6  \\
             & \textbf{ORPO} & 0.06 & - & - & -  \\
            \midrule
            \multirow{5}{*}{LoRA} 
             & \textbf{DPO} & 0.1 & - & - & -  \\
             & \textbf{SimPO} & 1.5 & 2.0 & - & -  \\
             & \textbf{IPO}  & 0.2 & - & - & -  \\
             & \textbf{KTO} & 0.5 & - & 1.0 & 0.3  \\
             & \textbf{ORPO}  & 0.02 & - & - & -  \\
            \bottomrule
        \end{tabular}
    }
    \caption{Hyper-parameter settings of other preference optimization methods.}
    \label{tab:hyper_beta}
\end{table}

\section{Case Study}
Table \ref{tab:case_study} presents a dialogue example from the ESC-Pro dataset. We show one round of dialogue with the prefer and non-prefer responses, along with the contextual background of the conversation, including the scene and description from the seed data, as well as the previous conversation history. By comparing the responses, we can see that the prefer response corresponds to the highest-quality strategy and reply, while the non-prefer responses are based on other strategies.

In Table \ref{tab:emotional_support}, we provide a case study to demonstrate that CSO can implicitly preserve emotional continuity over multiple turns by selecting context-appropriate strategies. This case shows that CSO can implicitly maintain emotional consistency across turns by adapting strategies to the seeker's changing emotions. The supporter begins with validation and empathy, then gradually introduces more proactive strategies like planning and reframing as the seeker becomes more hopeful. This smooth progression show that CSO responds not just to the current emotion, but also supports the emotional flow of the conversation as a whole.

\begin{table*}[!t]
\centering
\small
\setlength{\extrarowheight}{0pt}
\resizebox{\linewidth}{!}{
\begin{tabularx}{\textwidth}{l|X}
\toprule
\multicolumn{2}{l}{\textbf{Conversation Background}} \\
\midrule
\textbf{Scene} & Breakups or Divorce \\ \\
\textbf{Description} & After five years of marriage, my husband and I have decided to separate due to irreconcilable differences. I am feeling overwhelmed and uncertain as I navigate through this difficult time. \\
\midrule
\multicolumn{2}{l}{\textbf{Conversation History}} \\
\midrule
\textbf{Seeker} & Hey, I really need someone to talk to right now. Can you lend me an ear? \\ \\
\textbf{Supporter} & (Emotional Validation) Of course! I'm here for you. What's been going on? \\ \\
\textbf{Seeker} & My husband and I have decided to separate. It's been really hard to accept this reality. \\ \\
\textbf{Supporter} & (Emotional Validation) I'm so sorry to hear that. Separating from a long-term partner can be incredibly challenging. It's normal to feel overwhelmed and unsure about what lies ahead. Can you share more about your feelings? \\ \\
\textbf{Seeker} & I feel like everything fell apart so quickly. We had dreams about raising a family together and now it's all shattered. \\ \\
\textbf{Supporter} & (Emotional Validation) It's really tough when our dreams and plans change unexpectedly. You're experiencing a whirlwind of emotions right now. It's completely understandable that you're feeling devastated and lost. Know that you're not alone in going through this difficult process. \\ \\
\textbf{Seeker} & Thank you. I can't help but blame myself for the breakdown of our marriage. I keep wondering if I did something wrong. \\ \\
\textbf{Supporter} & (Emotional Validation) It's natural to feel a sense of self-blame during such a challenging time. However, I want to emphasize that relationships involve the contributions of both partners, and it's rarely a reflection of just one person's actions. You are worthy of love and happiness, regardless of the outcome of this marriage. \\ \\
\textbf{Seeker} & I appreciate your kind words. It's just hard to let go of all the memories and years we spent together. I miss him so much. \\ \\
\textbf{Supporter} & (Empathetic Statements) It's completely understandable that you're missing your husband. Remember that healing takes time and it's okay to grieve the loss of your marriage. It's important to be patient with yourself and allow yourself to process the many emotions that may arise. Lean on your support system and engage in self-care activities that bring you comfort and joy. \\ \\
\textbf{Seeker} & I'll try my best. Right now, it feels impossible to imagine moving on. Do you have any advice on how to start the healing process? \\
\midrule
\multicolumn{2}{l}{\textbf{Responses}} \\
\midrule
\textbf{Preferred} & (Affirmation) While it may seem overwhelming now, know that healing and growth are possible. Focus on taking care of yourself both physically and emotionally. Engage in activities that bring you happiness and surround yourself with loved ones who uplift you. Consider seeking support from a therapist or counselor who can help guide you through this process. You are stronger than you realize, and I believe in your ability to navigate through this difficult time. \\ \\
\textbf{Non-Preferred} & (Collaborative Planning) Absolutely! Let’s break it down together. What’s one small step you think you could take this week to begin healing? \\ \\
\textbf{Non-Preferred} & (Avoid Judgment and Criticism) Absolutely! Let’s break it down together. What’s one small step you think you could take this week to begin healing? \\ \\
\textbf{Non-Preferred} & (Reframe Negative Thoughts) I get that moving on feels impossible right now. But think about it this way: healing is a journey, not a race. Each small step counts. \\ \\
\textbf{Non-Preferred} & (Share Information) Starting the healing process can be tough, but small steps can help. Journaling your feelings, talking to friends, or exploring new hobbies can be beneficial. \\
\bottomrule
\end{tabularx}}
\caption{An example of preference dialogue in our ESC-Pro.}
\label{tab:case_study}
\end{table*}

\section{Human Evaluation Details}
\label{app:human_eval}

We conducted human evaluations and outsourced the evaluation tasks to anonymous crowd workers. We employed three evaluators for the human evaluation process, all of whom had prior annotation experience to ensure reliability. Before conducting evaluations, they received targeted training to thoroughly understand the criteria for each metric, promoting consistency and accuracy in their judgments. Evaluators were compensated based on the number of annotated samples they completed. To assess annotation consistency, we report the inter-annotator agreement using Cohen's Kappa scores, which are provided in the final column of Tables \ref{tab:preference_pair_comparison} and \ref{tab:human_evaluation_results}. Detailed guidelines and instructions were provided for each rating criterion to ensure the accuracy and consistency of the evaluations, as outlined in appendix \ref{app:data_quality}. For each evaluation task, we required three evaluators to assess 100 samples based on the corresponding metrics.

\begin{table*}[ht]
\centering
\renewcommand{\arraystretch}{1.5}
\begin{tabular}{l|l|p{10cm}}
\toprule
\multicolumn{3}{l}{\textbf{Conversation Background}} \\
\midrule
\textbf{Scene} & \multicolumn{2}{l}{Depression and Low Mood} \\
\textbf{Description} & \multicolumn{2}{p{15cm}}{I've been feeling emotionally exhausted and overwhelmed by everything happening in my life. It's hard to stay positive when it feels like things keep piling up and nothing changes.} \\
\midrule
\textbf{Role} & \textbf{Emotion / Strategy} & \textbf{Utterance} \\
\midrule
Seeker & Sadness & I've been feeling really overwhelmed lately. I don’t know how to handle everything that's going on. \\
Supporter & Emotional Validation & It sounds like you're carrying a lot right now, and it's understandable that you're feeling overwhelmed. It's okay to feel that way. \\
Seeker & Frustration & Yeah, it's just so much, and I feel like I can't get a grip on it. \\
Supporter & Empathetic Statements & It's tough when everything seems like it's piling up. I can hear how much you're struggling with it. \\
Seeker & Hopelessness & I'm trying to stay positive, but it's hard. \\
Supporter & Affirmation & It's really impressive that you're trying to stay positive despite everything. That shows real strength. \\
Seeker & Overwhelm & I just don’t know where to start. \\
Supporter & Collaborative Planning & Let's take it step by step. What’s one small thing you could focus on right now? \\
Seeker & Hope & Maybe I can tackle some smaller tasks to get the ball rolling. \\
Supporter & Provide Different Perspectives & That's a great perspective. Starting small can help you feel more in control and build momentum. \\
Seeker & Defeat & I feel like I’ve been stuck in this cycle, and nothing changes. \\
Supporter & Reframe Negative Thoughts & It's easy to feel stuck when it seems like things aren’t improving. But every small step forward counts, even if it doesn’t feel like it. \\
Seeker & Acceptance & You're right. Maybe I need to be kinder to myself through this. \\
Supporter & Avoid Judgment and Criticism & Absolutely. Being kind to yourself is key. You're doing your best, and that's enough. \\
Seeker & Hopeful & I think I can manage it now, one step at a time. \\
Supporter & Emotional Validation & That's a wonderful mindset. Taking it one step at a time will help you regain your balance. \\
\bottomrule
\end{tabular}
\caption{Dialogue turns with emotional Context and support strategies}
\label{tab:emotional_support}
\end{table*}

\end{document}